**Machine interpreting competitiveness: a discourse-sensitive comparison of meaning and target-language quality in English–Chinese simultaneous interpreting**

Yuanyuan Ding

Associate Professor, Middlebury Institute of International Studies at Monterey

Correspondence: yuanyuand@middlebury.edu; cherrydyy@gmail.com

**Abstract**

Machine interpreting (MI), or AI-enabled speech translation, is increasingly visible in professional and public-facing communication, yet its performance relative to human simultaneous interpreters remains insufficiently understood. Existing studies are often limited by a single discourse mode and evaluation frameworks that are not well aligned with interpreting as a discourse-level communicative activity. The present study compares five commercial MI systems with 26 human interpreters (14 professionals, 12 advanced students) on English-to-Chinese simultaneous interpreting across planned and unplanned discourse, with a particular focus on meaning and target-language quality. In these materials, MI performance differed sharply between the two discourse modes. In the planned formal speech, the strongest system became highly competitive on meaning, entering the upper human range, and a second system also performed strongly. In the extemporaneous interview, however, no MI system approached the stronger human tiers. Code-level analysis further shows that similar overall scores often masked different error profiles. Human interpreters were more likely to preserve the broader communicative trajectory through selective omission, restructuring, and contextual inference, whereas even the best MI systems were more vulnerable to semantic distortion, discourse-tracking failures, and context-insensitive interpretation. On target-language quality, the human advantage proved more stable than on meaning. Unintelligible or nonsensical output and disfluency carryover emerge as particularly salient weaknesses in machine performance, especially in spontaneous discourse. Overall, the findings are consistent with a discourse-sensitive account of where MI appears competitive, where it remains limited, and why these distinctions matter for interpreting practice, interpreter education, future MI development, and the broader deployment of AI-enabled speech translation across cross-linguistic communication settings.

**Keywords:** AI-enabled speech translation, machine interpreting, simultaneous interpreting, human–machine comparison, discourse mode, interpreting quality, meaning accuracy, target-language quality, error analysis, English–Chinese interpreting,

## 1 Introduction

Until recently, automated speech translation was used mainly in well-defined, fairly predictable situations, such as healthcare, travel and business (Horváth, 2021). Interpreting platforms are now building machine interpreting (MI) into their services, for settings that range from international conferences to community interpreting (Matsushita, 2026). Most commercial systems have been built on a cascade architecture, in which automatic speech recognition (ASR) feeds into neural machine translation (Fantinuoli & Prandi, 2021), although end-to-end models are now reaching the market (ByteDance, 2025). These systems are no longer experimental prototypes but commercially deployed products. This has raised the practical and scholarly question of how current MI systems compare with human simultaneous interpreters.

While a substantial literature compares human translators with text-to-text machine translation, research directly comparing human interpreters with real MI systems is considerably thinner. Moreover, several human–machine comparisons use text-to-text MT applied to transcripts of the source speech, rather than real MI systems. Lee and Cha (2023) present theirs as a comparison of human interpreting with machine translation. Liu and Liang (2024) and Ioannidis (2026) treat text-to-text MT as a proxy for MI and say so explicitly. This creates a fundamental input asymmetry: human interpreters work from incremental, real-time input under processing constraints, whereas the MT systems work from written transcripts, free of those constraints. Differences observed under such conditions may reflect the input disparity rather than genuine differences in interpreting capacity.

Studies that use real MI systems provide more ecologically valid comparisons. Fantinuoli and Prandi (2021), the closest precedent, used five 2-minute speech extracts, each interpreted by a different interpreter. They noted that their small corpus did not allow the observed trends to be generalized. Cheng et al. (2024) report human parity for an English–Chinese system on largely unprepared speech. Their human level, however, is a threshold derived from the literature, and no interpreters worked on the same speeches. Pan and Seeber (2026) compared four machine systems with professional interpreters' renditions of the same material in five target languages, using automatic metrics calibrated on expert ratings of the human renditions. Their materials were re-enacted conference scenes, which the authors describe as a single register, "prepared conference roundtable" (p. 550), and the machine outputs received no expert quality ratings. More recent studies (Matsushita, 2026; Korybski et al., 2026a; W. Zhang et al., 2025) have begun to address these gaps.

A further question is whether the result of such comparisons depends on the kind of speech being interpreted. The distinction between planned and unplanned discourse is well-established in linguistics, traceable to Ochs (1979), who defines unplanned discourse as "discourse that lacks forethought and organizational preparation" (p. 55) and planned discourse as "discourse that has been thought out and organized (designed) prior to its expression" (p. 55). Ochs explicitly frames the distinction as a continuum rather than a binary: most everyday communication is "relatively unplanned or relatively planned" (p. 55) rather than at either extreme. Her own examples of the two ends, "presidential addresses to the nation" and "spontaneous dialogues and multi-party conversations" (p. 53), closely parallel the source materials used in the present study (a prepared prime-ministerial address and an extemporaneous interview). Within interpreting studies, Gile (2009, pp. 192–200) treats delivery speed, information density, logical structure, and prosody among the factors that interact in determining interpreting difficulty. He also notes that prepared speeches read aloud are generally denser than spontaneous speech (p. 193). This offers a cognitive-load account of the differences that Ochs places along the planned–unplanned continuum.

To the author's knowledge, no existing study has systematically compared human and machine interpreting across discourse modes. There are good theoretical reasons to doubt that MI competitiveness holds steady from prepared to spontaneous speech. Human interpreters infer the speaker's meaning as the speech unfolds, drawing on context and their own knowledge (Setton, 2005). A cascade MI system instead works in separate steps: it turns speech into text, translates the text, and then turns the translation back into speech (Pöchhacker, 2022). Human interpreters also adapt dynamically through inferencing, restructuring, and audience-oriented mediation. These capacities are rooted in the situated cognition that distinguishes interpreting as a human activity (Pöchhacker, 2024). Agency, the capacity for real-time professional judgement, including error recovery and context-sensitive adaptation, also separates the two. Pöchhacker (2024) places fully automatic translation low on this dimension, and Sperber et al. (2025) note that choosing error-recovery strategies to suit the circumstances is currently achieved only by human interpreters. MI systems, by contrast, may appear stable precisely because they lack such adaptive agency. Fantinuoli (2026) argues on conceptual grounds that current MI systems are "optimised for a single mode of delivery" (p. 5) and cannot adjust their strategy when a speaker reads from a prepared text or changes pace or register.

The comparison also depends on how interpreting output is evaluated. Interpreting is a real-time, discourse-level communicative activity carried out under temporal pressure, incomplete input, and shifting contextual demands (Pöchhacker, 2016). Human interpreters regularly use tactics suited to real-time work. Segmentation and passivization reduce delay, while generalization and omission reduce the load on working memory (He et al., 2016). This selectivity under load has empirical support. Liu et al. (2004) found that professional interpreters outperformed students even though their general working-memory capacity was no greater. The authors attributed the difference, at least in part, to domain-specific skill, notably "the ability to select more important ideas from the speech input" under demanding task conditions (p. 19). Evaluation paradigms transferred from offline machine translation penalise these very strategies: reference-based metrics such as BLEU treat them as deviations rather than communicative choices (Paulik & Waibel, 2008; Korybski et al., 2026b). Neural MT metrics correlate fairly well with human judgements overall, but Wein et al. (2024) found that one top-ranked metric, MetricX, gave lower scores the more an output was summarised. They argue that summarisation should not be penalised where no meaning is lost. Pan and Seeber (2026) report a similar problem for interpreting itself: their machine systems showed higher median measured coverage of source segments than professional interpreters, but the authors caution that this measure can count professional compression as omission. LLM-based evaluation does not yet close this gap: in a direct comparison with trained human evaluators, ChatGPT-4o did not reproduce human error annotation consistently (Korybski et al., 2026b). Interpreting-specific evaluation instruments, applied by expert human raters, are therefore needed.

Raw error counts are also misleading when error types differ systematically. Ioannidis (2026), using a text-to-text MT proxy for English–Greek institutional speeches, found that the human interpreter made more errors in total than the AI system (49 vs. 28), mostly omissions. Yet the rate of major errors was similar (1.9 vs. 1.4 per 1,000 words). The two also differed in kind: the interpreter's major errors mainly lost information, whereas the AI's affected institutional reference and pragmatic stance. Similarly, Korybski et al. (2026a) reported unweighted error counts based on Barik's (1971) taxonomy in their Polish–English comparison, noting qualitatively that error types differed across human and machine output but without incorporating severity distinctions into the quantitative comparison. A properly designed evaluation must therefore distinguish strategic omission from information loss, and human adaptive strategies from machine-specific technical failures (Wein et al., 2024; Ioannidis, 2026). In a cascade system, a word misrecognised at the speech-recognition stage can carry through into the translation (Pöchhacker, 2024; Ioannidis, 2026). Such ASR errors are among the meaning errors found in real MI output (Korybski et al., 2026a).

Before hybrid human–machine configurations can be seriously assessed (Matsushita, 2026; Cela Gutiérrez, 2025), a clearer empirical picture of how human and machine performance diverge under different conditions is needed. The present study addresses these gaps by comparing five commercial MI systems with human interpreters at two expertise levels on English-to-Chinese simultaneous interpreting across two discourse modes: formal prepared speeches and extemporaneous interviews. The focus on English-to-Chinese adds a typologically important dimension, as English-to-Chinese SI requires structural reordering that is cognitively demanding for both humans and neural MT systems. Evaluation focuses on meaning accuracy and target-language quality, because these two dimensions can be assessed on equivalent terms across human and machine output. Both are measured with interpretation-specific instruments rather than transferred MT metrics. By combining aggregate performance comparison with fine-grained error-level analysis across discourse modes, the study examines not only whether MI approaches human performance, but under what conditions it does so, where the human advantage remains most pronounced, and what the resulting patterns imply for interpreting practice, interpreter education, and future MI development.

## 2 Method

### 2.1 Research Questions

This study addresses the following research questions (all concerning English-to-Chinese simultaneous interpreting):

**RQ1**: How do contemporary state-of-the-art machine interpreting (MI) systems compare with human interpreters in terms of meaning and target-language quality?

**RQ2**: To what extent do these relative performances vary across discourse modes (planned vs. unplanned)?

**RQ3**: What qualitative differences exist between MI systems and human interpreters in terms of strengths and limitations at a granular level?

Throughout, competitiveness refers to performance on these two dimensions relative to the human outputs on the same text. It does not cover latency, delivery, or the way the systems rewrote their on-screen text while the speech was in progress. Meaning was scored for its likely effect on a listener, as judged by expert raters (§2.5.4), but comprehension was not tested with real listeners.

### 2.2 Machine Interpreting Systems

Five MI systems were selected for evaluation: **Tencent, Microsoft Teams, Youdao, iFlyrec, and Doubao** (the first two being the speech translation functions embedded in Tencent Meeting and Microsoft Teams, respectively). In the Results and Discussion, the systems are referred to as Systems A to E. The labels were assigned at random and do not follow the order in which the products are listed above. The aim of the study is to compare machine and human interpreting across discourse modes, not to rank products, and the tested versions date from February 2026. Selection was based on three criteria: (1) prominence in the English–Chinese simultaneous interpreting market, as identified through consultation with interpreting agencies and end users; (2) availability as publicly accessible, user-facing simultaneous interpreting products rather than developer-oriented APIs or enterprise-contracted services; and (3) capacity to generate real-time extractable bilingual output (source-language ASR transcripts and target-language translations) suitable for systematic evaluation. Several other prominent systems (e.g., Google's real-time translation API, Meta's SeamlessM4T, OpenAI's real-time speech API, Wordly, KUDO) were considered but excluded because they did not meet these criteria (see Supplementary Methods for details). All five systems displayed real-time source-language transcripts alongside target-language translations. This is consistent with the cascade architecture (ASR → MT) that Fantinuoli and Prandi (2021) describe as the mainstream approach, but it does not establish the architecture of each product: end-to-end models can also output source-language text, and Doubao's underlying model may be of this kind (ByteDance, 2025).

**2.3 Dataset: Source Materials and Human Interpreting Data**

Both the source materials and the human interpreting performances were drawn from an experimental study investigating the impact of ASR assistance on English-to-Chinese simultaneous interpreting performance (Ding et al., under review). The human interpreting performances were collected during the week of February 9, 2026. The machine interpreting outputs were generated on February 26–27, 2026, using the publicly available versions of each system at that time.

The human interpreting dataset includes two primary cohorts (total n = 26): professional interpreters (n = 14; years of professional interpreting experience: M = 10.4, SD = 5.3, Mdn = 10, range = 3–20) and advanced students (n = 12), all with Chinese as their first language and English as their second language. Students were drawn from the graduate Translation and Interpretation program at the Middlebury Institute of International Studies at Monterey and had passed a formal gatekeeping procedure ensuring a meaningful floor of interpreting competence. Advanced students were included as a second human benchmark. Korybski et al. (2026a) made a similar three-way comparison of professional interpreters, final-year interpreting students and MI systems, and found that professionals outperformed students, who in turn outperformed MI. Each participant interpreted one text in each discourse mode with ASR support and the other without, in a counterbalanced within-subjects design with two stimulus versions (Ding et al., under review). On each text, half the outputs in each cohort were therefore produced with ASR. For the main analysis, ASR-assisted and unassisted performances are combined within each cohort, because tool-assisted interpreting remains human interpreting in which the interpreter retains full agency and decision-making authority. A sensitivity analysis shows that the main pattern holds against unassisted and ASR-assisted outputs separately (§3.1.3; see Supplementary Methods for the full pooling rationale). Each of the four source texts was interpreted by all 26 human participants, ensuring within-text comparability. This addresses a limitation of Fantinuoli and Prandi (2021), who had one interpreter per text and noted that their small corpus did not allow generalization.

Four source texts were drawn from two recordings representing two discourse modes. Planned discourse: a formal prepared speech delivered by former UK Prime Minister Rishi Sunak (October 26, 2023) on AI opportunities and safety (approximately 15 minutes; 2,122 words; ~140 words per minute, after a 10% speed reduction applied in the original study to approximate typical interpreting delivery rates). Unplanned discourse: an extemporaneous interview with Anthropic CEO Dario Amodei at the World Economic Forum (Wall Street Journal, January 2026) on AI opportunities and risks (approximately 14 minutes; 2,344 words; ~170 words per minute, after the same 10% speed reduction). Although the unplanned-discourse rate is nominally higher, both rates fall within typical ranges for their respective discourse modes, and the unplanned segments' lower informational density may partly offset the higher rate (see Supplementary Methods, Source Material Descriptors). Both recordings feature native English speakers with standard accents. Neither source contained dense specialized terminology, reducing the role of domain-specific preparation in performance outcomes. From each source, two consecutive segments of approximately equal length were selected, yielding four source texts: Texts 1–2 (planned) and Texts 3–4 (unplanned). "Text" is used here for a stretch of spoken source material. All four were heard as recorded speech.

### 2.4 Design and Procedure

The study adopts an exploratory, descriptive design. Descriptive statistics and qualitative error analysis are used to map performance patterns across conditions; inferential statistical tests are not applied to group-level MEEC comparisons, as the sample sizes are not designed to support such analyses. Limited inferential tests are reported only where they directly support specific subsidiary findings: rater reliability (§2.5.2), the cohort × cross-mode performance pattern (§3.4.2), and non-parametric robustness checks (Supplementary Statistical Analysis).

Throughout the analysis, particular attention is given to the best-performing MI systems rather than to the machine-group mean. This framing reflects two practical realities. First, users deploy the strongest available system, not an average across all options, so the upper bound of MI performance is the more ecologically relevant comparison point. Second, unlike human interpreters, who are subject to fatigue, availability constraints, and finite concurrent capacity, a single high-performing MI system can be scaled to serve an essentially unlimited number of simultaneous sessions; the best system's performance therefore represents a deployable reality at scale.

Human interpreters were informed of the topic only upon arrival and received no advance preparation materials. Source audio was routed directly through system audio into each MI system to ensure identical input conditions across systems and eliminate environmental variability.[i] Each system was run once on each text, and its transcript was taken from the final text displayed on screen.

### 2.5 Assessment Method

#### 2.5.1 Conceptual Framework

Human performances were scored in the parent study (Ding et al., under review). The present study applies the same assessment framework, raters and codebook to the machine outputs (§2.5.5). Interpreting quality is widely conceptualised as a multidimensional construct spanning content fidelity, target-language acceptability, and delivery (Lee, 2008; Han, 2018; Pöchhacker, 2016). The present study focuses on meaning and target-language quality as its two assessment dimensions, capturing the core communicative function of interpreting: faithful transfer of source-language content in linguistically acceptable target-language form. These two dimensions reflect what conference delegates value. In Kurz's (1993) survey of three delegate groups, sense consistency with the original message, correct terminology and logical cohesion received the highest ratings. Correct grammatical usage and delivery features such as native accent and pleasant voice received the lowest. The delivery dimension (e.g., prosody, fluency, voice quality) was excluded because the majority of MI systems examined produce textual rather than audio output, and including it would introduce an asymmetric comparison.

Interpreting quality can be assessed from a product-oriented or a communicative perspective (Fantinuoli & Prandi, 2021). It can also be scored atomistically, by counting errors or content points, or holistically, by rating the performance as a whole (Han, 2018). The present study combines both perspectives and both scoring approaches. Meaning is scored atomistically, as a weighted count of errors, but each error is defined by its communicative impact: raters judged whether the key message reached a hypothetical listener, not merely whether information units were preserved (§2.5.4). This responds to a criticism of atomistic scoring that Han (2018) reviews: that it works at the lexical rather than the discourse level. This communicative orientation requires professional interpretive expertise, the kind of situated judgement that distinguishes expert assessment from mechanical error counting. Target-language quality is scored holistically, using band descriptors adapted from Lee (2008) (§2.5.3).

### 2.5.2 Rater Qualifications and Calibration

All human and machine outputs were evaluated by two expert raters with over 15 years of professional and pedagogical experience in conference interpreting. Both are senior practitioners in the English–Chinese conference-interpreting community. The principal investigator served as one of the two raters; the second rater was external to data collection (not involved in recruitment, session administration, or data handling). All outputs were scored from de-identified materials (human performances from coded audio, machine outputs from coded text), and the principal investigator did not consult the code-to-identifier mapping during scoring. Both raters had prior professional acquaintance with a subset of the 26 participants, so incidental recognition from voice or stylistic cues cannot be excluded. This and the investigator's dual role are discussed as limitations in §3.5.

To verify comparable individual standards, a calibration phase was conducted using benchmark samples, aligning on rubric descriptors, error classification, and severity thresholds. Independent pre-consensus ratings of five human outputs gave high point estimates of agreement: meaning accuracy (MEEC), ICC(A,1) = .970, 95% CI [.741, .997], $p = .001$; target-language quality (100-point scale), ICC(A,1) = .926, 95% CI [.270, .992], $p = .001$ (two-way mixed-effects, absolute agreement, single measures; see Supplementary Table S8). Because only five outputs were rated, the confidence intervals are wide. Koo and Li (2016) recommend judging reliability by the confidence interval rather than the point estimate. On that basis, agreement ranged from moderate to excellent for meaning accuracy and from poor to excellent for target-language quality. All subsequent evaluations used consensus coding: both raters worked jointly in ATLAS.ti, with disagreements resolved through discussion.

### 2.5.3 Target-Language Quality

Target-language quality was assessed using a rating scale adapted from Lee (2008), whose six-band descriptors for target-language production were mapped to a 100-point scoring system (band 6 = 90, band 5 = 80, band 4 = 70, band 3 = 60, band 2 = 50, band 1 = 40). The descriptors cover linguistic correctness, naturalness and contextual appropriateness. A seventh band (score 30, "Very poor") was added to accommodate machine outputs that, while technically complete, were largely unintelligible, a performance level not anticipated in the original human-oriented scale (see Supplementary Appendix B for the full rating scale). Each output received a single agreed score. In addition to the overall score, specific target-language quality features (awkwardness/register, disfluency carryover, and unintelligible or nonsensical output) were coded as part of the same consensus process.

### 2.5.4 Meaning

Meaning was assessed using a weighted-accuracy model based on discourse-level key-message segments. Each segment contains a core message and its supporting detail, weighted by communicative salience. This approach adapts the principle of weighted propositional analysis (Gieshoff & Albl-Mikasa, 2024) to a coarser granularity suited to cross-condition comparison at scale, because excessively fine-grained units risk directing rater attention toward surface-level detail at the expense of discourse-level coherence. For the present dataset, each source text yielded about 25 key-message segments. Segmentation was performed jointly by the two raters through consensus. A segment typically received at most one major code; a second was assigned only when an additional deviation was qualitatively distinct and independently damaged the key message (Supplementary Appendix A). Minor codes were not limited in number within a segment.

Errors were identified using a structured taxonomy and code book (Supplementary Appendix A) and classified as major or minor. Major errors are deviations that distort, contradict, or omit key message content, even if some supporting factual content is technically present in the output (3 codes: major meaning/logic error, major omission, major addition). Minor errors are deviations that affect precision without compromising the core message (7 codes: minor meaning shift, minor logic unclarity, minor addition, minor omission, minor list omission, minor number, minor terminology). Minor omission and minor list omission are separate codes: an omitted list item is coded as minor list omission only, so counts of minor omission exclude list omissions. A 5:1 weighting ratio was applied, whereby one major error is treated as equivalent to five minor errors, yielding the major-equivalent error count (MEEC): MEEC = (major errors) + (minor errors / 5). This ratio was empirically informed by informal consultation with ten practicing interpreters, in which respondents were asked to estimate the number of minor errors that, in their judgement, would be equivalent in communicative impact to one major error; the modal and median response was 5:1. Freitag et al. (2021) arrived at the same 5:1 ratio in a large-scale Multidimensional Quality Metrics (MQM) evaluation of machine translation. They tested major weights from 1 to 10 against a minor weight of 1, and a major weight of 5 gave the most stable system rankings. The main pattern was not sensitive to this choice. With major errors only, and with 3:1 and 10:1 weightings, System B outperformed 15 to 22 of the 26 human outputs on the planned texts and 3 to 6 on the unplanned texts (Supplementary Table S14). In addition, target-language quality codes were applied alongside meaning codes for qualitative profiling; neither entered the calculation of meaning accuracy scores. While the weighted error approach shares structural features with analytic translation quality frameworks such as MQM (Lommel et al., 2014), the present codebook was designed specifically for interpreting evaluation: each error is located within a discourse-level key-message segment, and its severity is judged by its effect on a listener's understanding of that segment's key message.

#### 2.5.5 Machine Output Evaluation and Blinding

Machine outputs were evaluated in a separate round using the same framework. System identities were removed and replaced with numeric codes. For each source text, the five machine outputs were mixed with five human transcripts from the dataset (20 in total). These were verbatim transcriptions of the de-identified human audio recordings, prepared for this round. The two raters evaluated each set of ten together. The round served two purposes.

First, it reduced the risk of bias. Because machine and human transcripts were intermixed, raters could not approach them as separate sets. Residual awareness of condition membership cannot be entirely ruled out, particularly for weaker-performing machine systems in unplanned discourse, but anonymised transcripts and randomised presentation order mitigated potential bias. If raters applied stricter standards to suspected machine output, MI systems' competitive performance on planned speech would be understated rather than overstated. The same bias could, however, widen the planned–unplanned contrast, because machine output was easier to recognise in unplanned discourse.

Second, it served as a check on evaluation format. In the main evaluation, human performances were scored primarily from audio, whereas machine outputs could be scored only from transcripts. The 20 human transcripts were therefore scored in the same three-column format as the machine outputs (source text, ASR output and Chinese output), and their scores were compared with the audio-based scores of the same performances. Target-language quality scores were virtually unchanged and MEEC scores were closely similar, indicating that the difference in evaluation format did not systematically shift scores (see Supplementary Methods, Blinding Procedures and Transcript-Based Evaluation Validity).

A supplementary cause-of-error code (ASR-caused error) was applied only to machine-interpreting outputs in a separate second-round review: after the initial blind evaluation was complete and system identities had been revealed, the same two raters revisited the major errors coded for machine outputs and compared each against the corresponding machine-generated English ASR transcript to determine whether the error originated in ASR failure (see Supplementary Appendix A for the full procedure). The judgement rests on what the two columns show: an error was coded as ASR-caused when the Chinese output followed an error visible in the displayed English transcript, such as a misrecognised word or a misplaced sentence boundary. This points to speech recognition as the source of the error. It cannot show that the displayed transcript was the text each system translated from, because the study observed only the on-screen output of each system and did not verify its internal design (§2.2). The ASR-caused error code did not enter the calculation of meaning accuracy scores.

**2.5.6 Analytical Approach**

Because only five MI systems were available for comparison, formal inferential tests comparing MI and human groups would have had very low statistical power and were therefore not conducted for the main analyses. The study instead relies on descriptive comparisons (MEEC distributions, code-level error profiles, and within-text rank positions) to characterise the human–machine performance landscape. Where inferential statistics are reported, group comparisons and Fisher's exact test are two-sided with $\alpha = .05$, and any one-sided result is labelled as such. ICC estimates are reported with 95% confidence intervals and the standard F test against zero.

## 3 Results and Discussion

### 3.1 Meaning: Aggregate Performance

The aggregate meaning results are organised by discourse mode. Each MI system is positioned against all 26 human outputs, and mean scores are also reported separately for professional interpreters (n = 14) and advanced students (n = 12).

#### 3.1.1 Planned Discourse (Texts 1 and 2)

The planned discourse pair exhibits a clear, internally consistent pattern. In both texts, the professional cohort outperformed the advanced-student cohort on average, and the strongest MI systems occupied positions much closer to the human benchmark than the weaker systems. Strong MI performance appeared in both texts, which are consecutive segments of the same speech.

System B was the strongest MI system, scoring 7.0 major-equivalent errors on both texts and outperforming 22 and 21 of the 26 human outputs, respectively, which placed it consistently within the upper human range. System E also performed strongly (8.8 and 10.8), entering the stronger part of the human distribution on both texts. System C (10.4 and 15.8) remained within the overall human range but fell further from the strongest human tier, particularly on Text 2. System D and System A fell below all human performances on both texts (Table 1, Figure 1).

Taken together, the two planned texts show that MI is not a homogeneous category. The gap between the strongest and weakest systems is large, but the key substantive point is the stability of the upper bound: System B, and to a lesser extent System E, remained highly competitive with human performance across both planned texts. This convergence should not be overgeneralised: it reflects upper-bound MI performance under planned, read-out conditions rather than general equivalence with professional simultaneous interpreting.

***Table 1**. Position of individual MI systems on the two planned discourse texts (major-equivalent error counts (MEEC); lower = better)*

| System | Text 1 (MEEC; outperformed) | Text 2 (MEEC; outperformed) | Position relative to pooled human benchmark |
|---|---|---|---|
| System B | 7.0 (22/26) | 7.0 (21/26) | Consistently within the upper human range |
| System E | 8.8 (19/26) | 10.8 (16/26) | Within or near the upper human range |
| System C | 10.4 (18/26) | 15.8 (9/26) | Within the overall human distribution but below the strongest human tier |
| System D | 26.2 (0/26) | 28.0 (0/26) | Outside the human range on both planned texts |
| System A | 29.8 (0/26) | 34.8 (0/26) | Outside the human range on both planned texts |

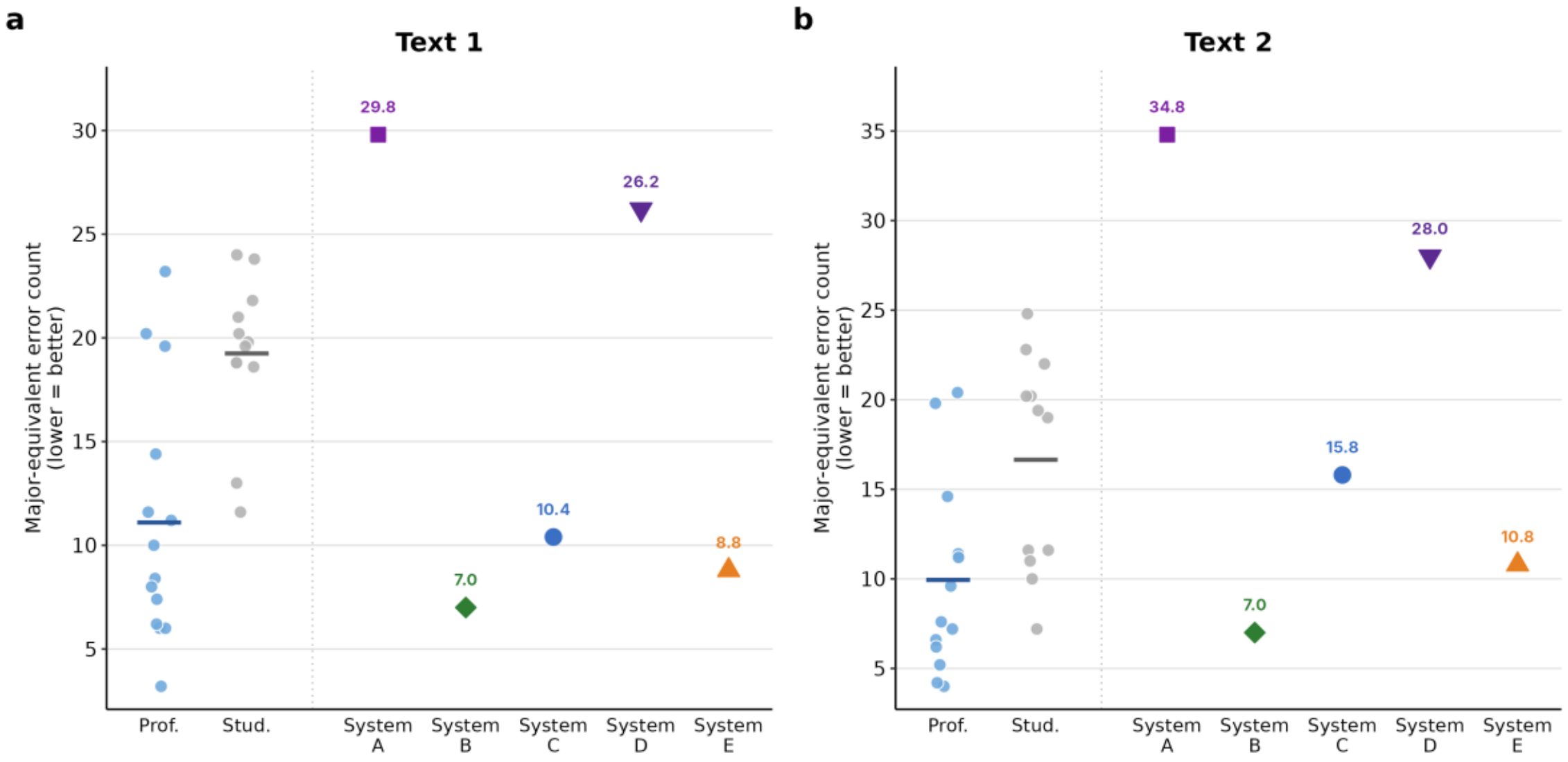

**Figure 1.** Planned discourse (Texts 1 and 2): individual MEEC scores for professional interpreters (blue dots, n = 14), advanced students (grey dots, n = 12), and MI systems (labelled markers). Horizontal lines indicate group means for professionals and students. Panel (a) shows Text 1; panel (b) shows Text 2.

### 3.1.2 Unplanned Discourse (Texts 3 and 4)

The unplanned discourse pair produced a markedly different configuration. Across both texts, the upper bound of MI performance shifted downward relative to the human benchmark. Unlike in the planned discourse, no MI system consistently approached the stronger human tier.

In Text 3, the professional cohort averaged 7.94 major-equivalent errors, compared with 11.78 for the student cohort. Against this benchmark, System B (14.4) was the strongest MI system but outperformed only 3 of 26 human outputs. System C (15.0) and System E (15.6) were close to one another and each outperformed only 2. System D and System A, both at 20.8, fell below all human performances.

Text 4 showed the same general discourse-mode effect, though somewhat less severely for the strongest systems. The professional cohort averaged 9.14 major-equivalent errors, compared with 15.73 for the student cohort. System E (14.4) was the strongest MI system and outperformed 10 of 26 human outputs, while System B (17.4) and System C (18.2) outperformed 6 and 4, respectively. System D (23.4) and System A (22.4) again fell outside the human range (Table 2, Figure 2).

These results suggest that current MI systems remain much more vulnerable than human interpreters to spontaneous, context-dependent discourse. Although Korybski et al. (2026a) did not formally analyse the discourse-mode contrast as a variable, one of their examples illustrates a similar vulnerability to features of improvised speech: when the speaker self-corrected mid-utterance, accredited interpreters exploited the moment for effective on-the-fly editing, whereas MI providers produced incoherent or drastically reduced output. Fantinuoli and Prandi's (2021) results differ on this point. Their aggregate results, drawn predominantly from impromptu speeches (four of five), showed near-parity on informativeness, whereas the present data reveal a widening meaning gap in extemporaneous discourse. This divergence may reflect differences in language pair, MI system generation, or the nature of the unplanned material, and underscores the value of disaggregating results by discourse mode.

***Table 2****. Position of individual MI systems on the two unplanned discourse texts (major-equivalent error counts (MEEC); lower = better)*

| System | Text 3 (MEEC; outperformed) | Text 4 (MEEC; outperformed) | Position relative to pooled human benchmark |
|---|---|---|---|
| System B | 14.4 (3/26) | 17.4 (6/26) | Near the lower end of the human distribution |
| System C | 15.0 (2/26) | 18.2 (4/26) | Near the weaker end of the human distribution |
| System E | 15.6 (2/26) | 14.4 (10/26) | Most competitive on Text 4, but still below the stronger human tiers |
| System D | 20.8 (0/26) | 23.4 (0/26) | Outside or below the human range |
| System A | 20.8 (0/26) | 22.4 (0/26) | Outside or below the human range |

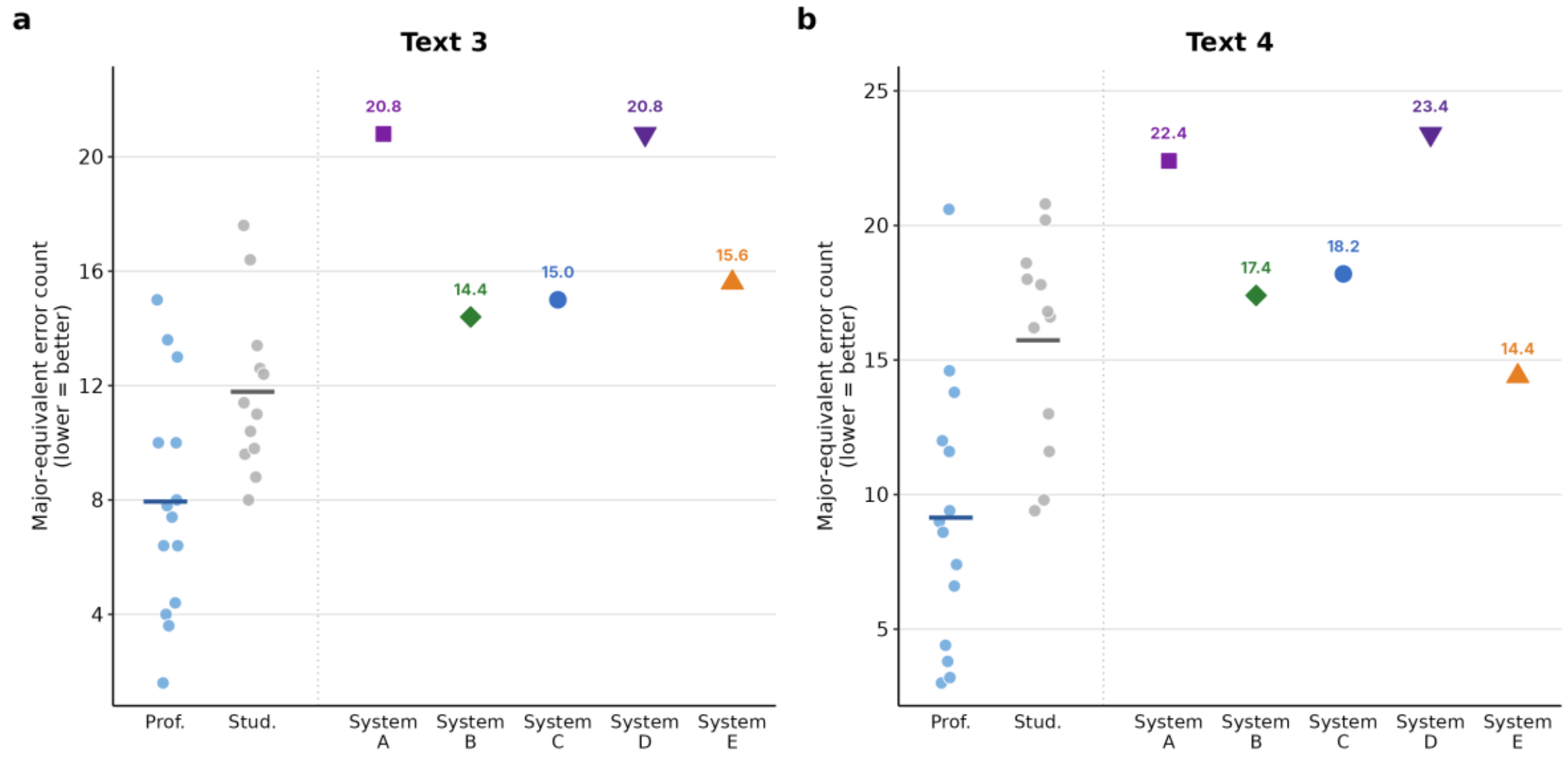

**Figure 2.** Unplanned discourse (Texts 3 and 4): individual MEEC scores for professional interpreters (blue dots, n = 14), advanced students (grey dots, n = 12), and MI systems (labelled markers). Horizontal lines indicate group means for professionals and students. Panel (a) shows Text 3; panel (b) shows Text 4.

#### 3.1.3 Cross-Mode Synthesis

Taken together, the aggregate meaning results show that the relative standing of MI systems is strongly discourse-sensitive in these materials, so there is no single, discourse-independent answer to the question of where MI systems "stand." Within-text rank shifts illustrate this. MI systems' mean rank dropped by 7.8 positions from planned to unplanned texts. The shift was most pronounced for the competitive systems: System B's mean within-text rank moved from 5.5 to 23.0, System E from 10.5 to 22.0, and System C from 15.5 to 25.2. Given that only five MI systems were available per text, formal inferential comparisons were not conducted; the analysis instead relies on descriptive comparisons and code-level error profiles, which are better suited to the exploratory design and small system-level samples.

This pattern does not depend on combining ASR-assisted and unassisted human outputs. When each MI system was compared separately with the 13 unassisted and the 13 ASR-assisted outputs on each text, the planned–unplanned contrast held in both subsets (Supplementary Table S13). System B outperformed between 9 and 13 of the 13 human outputs in each subset on the planned texts, and between 0 and 5 on the unplanned texts. Because each subset contains only 13 outputs, these counts are descriptive.

### 3.2 Meaning: Code-Level Profiles in Representative Texts

To complement the aggregate comparison, the following code-level analysis focuses on one representative text from each discourse mode: Text 2 for planned discourse and Text 3 for unplanned discourse.

#### 3.2.1 Planned Discourse (Text 2)

The formal prepared speech shows partial convergence in weighted totals but not in code composition. System B's MEEC (7.0) placed it within the same broad performance zone as the eight strongest human outputs (MEEC range 4.0–7.6; mean = 6.0), yet Table 3 and Figure 3 show that this convergence masked substantially different error profiles. A natural performance gap separates these eight interpreters from the next-ranked human (MEEC = 9.6), making this cluster a well-defined comparison group rather than an arbitrary selection.

The System B–human comparison is especially revealing. Relative to the eight strongest human outputs as a group (mean MEEC = 6.0), System B produced more major meaning/logic errors (4 vs. 1.2) and substantially more minor meaning shifts (9 vs. 2.4), but far fewer minor omissions (0 vs. 5.9) and no minor list omissions (0 vs. 4.5). Distortion was the predominant error type for MI, whereas omissions were more characteristic of human output.

This divergence suggests that human and machine errors arise in different ways under dense prepared-speech conditions. Stronger human interpreters appear more willing to leave lower-priority details and list items under pressure, thereby protecting the main message at the cost of local completeness. MI systems, by contrast, preserve more surface material but remain more vulnerable to semantically incorrect transfer, especially when ASR or downstream alignment fails. For example, in the planned discourse line "Just think for a moment about what that will mean for our country. The growth it will catalyse, the jobs it will create, the change it can deliver—for the better," human interpreters often omitted one or more elements in the parallel structure, whereas the strongest MI systems generally preserved all three. This helps explain why top MI systems are most competitive in the planned texts: when discourse is syntactically clear, rhetorically parallel, and lexically transparent, machines can retain surface completeness with a consistency that human interpreters do not always maintain under real-time pressure.

This distortion-versus-omission contrast converges with Fantinuoli and Prandi (2021), who found that human interpreters outperformed MI on intelligibility while machines were marginally stronger on informativeness, and with He et al. (2016), who found that interpreters omit more source content than translators. They traced this to the limits of human working memory, a constraint that machines do not share.

Another notable pattern observed from the code profile concerns the dependence of machine major errors on ASR failure. In the post-hoc second-round review (see §2.5.5), raters compared each major error coded for machine outputs against the corresponding ASR transcript. System B showed a high ASR-caused burden: 4 of its 5 major errors (80.0%) were judged to have originated in ASR failure. The proportions ranged from 62.5% to 87.9% across all five MI systems. In the formal speech, the main bottleneck for strong MI performance still appears to lie upstream in speech recognition accuracy.

***Table 3***. *System B versus the top eight human outputs on selected meaning-related codes in the formal prepared speech*

| Performance | MEEC | Major meaning/logic | Major omission | Minor meaning shift | Minor omission | Minor list omission |
|---|---|---|---|---|---|---|
| System B | 7.0 | 4 | 1 | 9 | 0 | 0 |
| Human 019 | 4.0 | 0 | 1 | 0 | 6 | 4 |
| Human 033 | 4.2 | 2 | 0 | 1 | 2 | 4 |
| Human 005 | 5.2 | 2 | 1 | 3 | 5 | 2 |
| Human 028 | 6.6 | 0 | 2 | 4 | 10 | 7 |
| Human 035 | 6.2 | 3 | 1 | 1 | 2 | 5 |
| Human 009 | 7.2 | 2 | 2 | 4 | 6 | 3 |
| Human 027 | 7.2 | 1 | 2 | 2 | 8 | 6 |
| Human 001 | 7.6 | 0 | 3 | 4 | 8 | 5 |

*Note.* MEEC = major-equivalent error count.

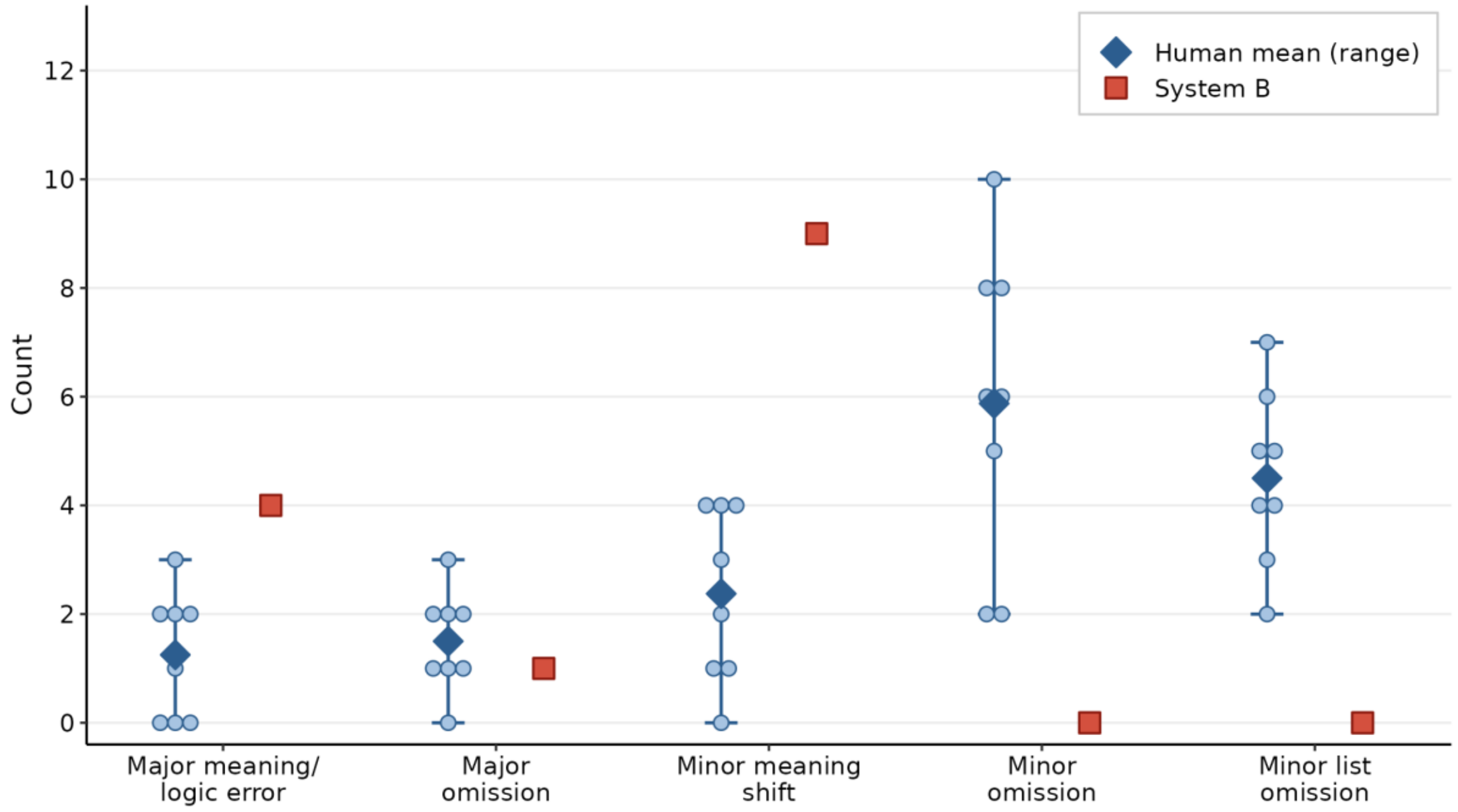

**Figure 3.** Formal prepared speech (Text 2): System B versus the top eight human outputs on selected meaning-related error codes. Each data point represents one output. Range bars represent the observed range (minimum to maximum) across the eight human outputs in the comparison group. The single System B value is shown without a range bar because it represents one system output.

### 3.2.2 Unplanned Discourse (Text 3)

The extemporaneous interview shows a sharper separation between human and machine performance. The four best human outputs range from 1.6 to 4.4 in MEEC, whereas the three strongest MI systems cluster at 14.4–15.6.

At the level of weighted totals, the three strongest MI systems (MEEC mean = 15.0) and the four nearest human outputs (MEEC mean = 14.60) are close. Yet Table 4 and Figure 4 show a pronounced divergence in code composition. The top three MI systems average 11.0 major meaning/logic errors, 7.67 minor meaning shifts, and 8.0 minor unclear-logic codes, compared with 5.75, 4.0, and 0.75, respectively, for the nearest humans. The human outputs show far more major omissions (5.50 vs. 0.33) and minor omissions (10.50 vs. 0.67).

The broader dataset reinforces this pattern. Across all five machines, mean counts exceed those of humans for major meaning/logic error (12.8 vs. 4.04), minor meaning shift (7.0 vs. 3.65), and minor unclear logic (6.0 vs. 1.88). Humans again show more major omission (3.08 vs. 0.8) and minor omission (6.92 vs. 0.4). Unlike the formal speech, minor list omission is nearly absent in both groups, suggesting that list density is not the main source of difficulty in this extemporaneous discourse.

The code-level picture, therefore, refines the discourse-sensitivity claim of §3.1. In the interview, the key machine weakness is not selective information loss but instability in meaning construction and discourse linkage. Human interpreters appear more able to preserve the broad communicative trajectory by selectively omitting information under pressure; MI systems are more prone to semantic distortion, weakly signaled logic, and local breakdown in output coherence once the discourse becomes spontaneous and less predictable.

The same second-round ASR-attribution procedure applied to the unplanned-discourse major errors yields a more differentiated picture. Unlike the formal speech, ASR dependence was low in the strongest systems: only 18.2% of System B's major errors and 27.3% of System C's were judged to have originated in ASR failure. By contrast, 94.4% of System D's and 83.3% of System A's major errors remained ASR-caused. This pattern suggests that speech recognition still holds back the weaker systems in spontaneous speech. For the stronger systems, the deeper limitation may lie in their inability to process contextual and discourse-level meaning beyond the sentence.

***Table 4***. *Top MI systems versus the four nearest human outputs on selected meaning-related codes in the extemporaneous interview*

| Performance | MEEC | Major meaning/logic | Major omission | Minor meaning shift | Minor omission | Minor unclear logic |
|---|---|---|---|---|---|---|
| System B | 14.4 | 10 | 1 | 7 | 1 | 7 |
| System C | 15.0 | 11 | 0 | 9 | 0 | 8 |
| System E | 15.6 | 12 | 0 | 7 | 1 | 9 |
| Human 010 | 13.6 | 4 | 7 | 4 | 7 | 0 |
| Human 003 | 13.4 | 7 | 3 | 3 | 13 | 1 |
| Human 037 | 15.0 | 8 | 4 | 5 | 9 | 1 |
| Human 002 | 16.4 | 4 | 8 | 4 | 13 | 1 |

*Note.* MEEC = major-equivalent error count.

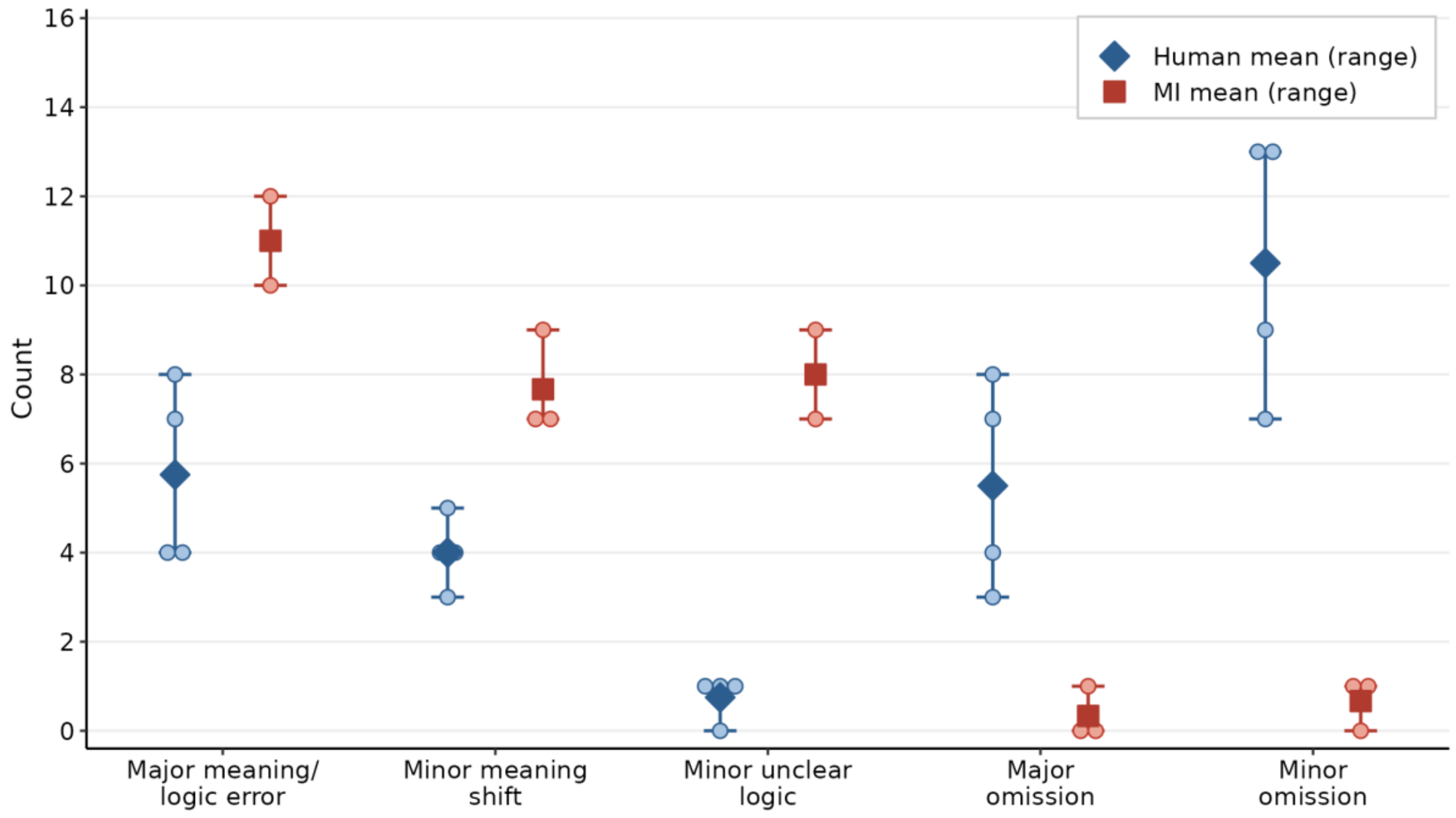

**Figure 4.** Extemporaneous interview (Text 3): top three MI systems versus the four nearest human outputs on selected meaning-related error codes. Each data point represents one output. Range bars represent the observed range (minimum to maximum) across individual outputs within each comparison group.

### 3.2.3 Recurrent Machine Error Patterns: Illustrative Examples

Four illustrative examples are presented below to exemplify recurrent patterns by which MI output deteriorates (see the Supplementary Note for additional examples).

***Example 1. Lexical misrecognition without contextual repair***

Source: “That’s not just about having the technical skills, but the raw computing power.”

System E ASR: “…the rule computing power.” → System E output: “这不仅仅关乎掌握技术技能，还涉及到有规则的计算能力。”

System B ASR: “…the real computing power.” → System B output: “这不仅关乎技术技能，更关乎真正的计算能力。”

Here the source adjective *raw* is misrecognised as *rule* and *real*. Both systems translate the misrecognised form compositionally, producing distorted output. A human interpreter uncertain about *raw* would be likely to preserve the broader contrast by rendering it more generally as 算力 or 计算能力. The example illustrates a central machine weakness: once an acoustically plausible but semantically inappropriate word is recognised, the system lacks the discourse-level judgement needed to suppress or neutralise it.

***Example 2. Speaker-turn collapse***

Source: Speaker 2: “and thank you for having me.” Speaker 1: “Not at all.”

System B ASR: “And thank you for having me not at all.” → System B output: “感谢没有让我亲自出席。”

System B treats two short turns by different speakers as a single utterance. The routine politeness exchange is reanalysed as one sentence with reversed meaning. No human interpreter in the dataset produced this type of error. The example underscores the importance of interactional parsing in spontaneous discourse. The concatenation was a consistent feature of the tested systems' output rather than an isolated error (see §3.4.3).

***Example 3. Plausible word form, implausible discourse sense***

Source: "…where the model is getting more and more cognitively capable every few months. And that that march has just been constant."

System B output: "而三月份一直保持着这种状态。" System C output: "而三月一直很稳定。"

Four of the five MI systems interpreted *march* as the calendar month March rather than as a metaphorical reference to continued progression. None of the human interpreters made this error. The example points to a downstream weakness: even when the recognised word is acoustically plausible, the system may fail to select the discourse-appropriate sense.

***Example 4. Misrecognition combined with implausible propositional construction***

Source: "I have some engineers, some engineering leads within Anthropic, who have basically said to me, 'I don't write any code anymore.'"

System E ASR: "I have some engineers, some engineering leaks within Anthropic who? Have basically said to me, I don't write any code anymore."

System E output: "我有一些工程师，在人工智能研究公司 Anthropic 内部存在一些工程信息泄露，是谁干的呢？基本上是跟我说，我不再写任何代码了。"

The ASR system mishears *leads* as *leaks* and, compounded by erroneous segmentation, generates an entirely new proposition about an engineering information leak and introduces a spurious question. A human interpreter might occasionally mishear *leads*, but would be unlikely to construct a proposition about data leakage in this context, because the continuation ("I don't write any code anymore") strongly constrains interpretation toward a personnel statement rather than a security incident. The example highlights human interpreters' reliance not only on acoustic decoding but also on pragmatic plausibility and world knowledge as real-time filters on candidate meanings.

### 3.3 Target-Language Quality

#### 3.3.1 Aggregate Target-Language-Quality Patterns Across the Four Texts

The aggregate results show that professional interpreters had the highest mean target-language quality score in all four texts. On the planned texts, the highest-scoring system compared less favourably with the human outputs on target-language quality than on meaning. The highest-scoring system, System E, scored higher than 11 of the 26 human outputs on Text 1. On Text 2 it scored higher than 11 and the same as a further 9. On meaning, by comparison, the strongest system outperformed 22 and 21 of the 26. Some machine outputs were locally strong and occasionally idiomatic, but overall, they remained less stable and less acceptable than the human outputs, especially those of the professional interpreters.

Target-language quality was also discourse-sensitive. In the two planned texts, the stronger MI systems performed noticeably better, with System E and System B producing the highest machine scores and at times generating concise and relatively natural Chinese output. In the two unplanned texts, machine scores declined, and no system outscored any human output, although System E scored the same as 7 of them on Text 3. This decline was not confined to one weak system or one isolated text; rather, it appeared across most systems and was sufficiently regular to suggest that extemporaneous discourse places heavier pressure on target-language quality than planned, read-out discourse.

MI is not a homogeneous category on this dimension either. System E was the strongest system overall, especially in the planned pair, while System B remained relatively competitive across all four texts. System C occupied an intermediate position, and System A was consistently weakest.

Even so, none of the systems matched the professional human benchmark across the four texts. The stronger MI systems were clearly capable of producing stretches of fluent, concise, and usable Chinese. What distinguished them from human outputs was not the total absence of good target-language production, but rather the lack of consistency across discourse conditions, a finding broadly consistent with the human advantage in intelligibility reported by Fantinuoli and Prandi (2021), whose intelligibility criterion is closest to the holistic score used here.

***Table 5**. Aggregate target-language-quality scores across the four texts (cohort means by text)*

| Cohort / System | Text 1 | Text 2 | Text 3 | Text 4 | Planned M (SD) | Unplanned M (SD) |
|---|---|---|---|---|---|---|
| Professionals (n = 14) | 72.1 (9.6) | 70.4 (8.9) | 73.2 (6.1) | 75.0 (5.5) | 71.2 (7.8) | 74.1 (5.2) |
| Advanced students (n = 12) | 65.0 (6.7) | 65.8 (7.9) | 65.0 (5.6) | 65.0 (6.7) | 65.4 (6.6) | 65.0 (4.9) |
| System C | 50.0 | 60.0 | 45.0 | 50.0 | 55.0 | 47.5 |
| System B | 60.0 | 65.0 | 55.0 | 55.0 | 62.5 | 55.0 |
| System E | 65.0 | 70.0 | 60.0 | 55.0 | 67.5 | 57.5 |
| System A | 40.0 | 40.0 | 30.0 | 40.0 | 40.0 | 35.0 |
| System D | 50.0 | 50.0 | 55.0 | 50.0 | 50.0 | 52.5 |

*Note.* Higher scores indicate better target-language quality. Per-text values for cohort rows show group means with standard deviations in parentheses (professionals n = 14; students n = 12). Machine rows show single consensus scores. Full individual-level distributions are reported in the Supplementary Data file.

### 3.3.2 Code-Level Target-Language-Quality Profiles in the Representative Texts

The three codes examined (awkwardness/register, disfluency carryover, and unintelligible or nonsensical output) behave differently across discourse modes.

In the planned discourse, disfluency carryover was not the main machine problem, as the source speech was clean and low in disfluency. The more diagnostic weaknesses were awkwardness/register and unintelligible or nonsensical output. The stronger systems could at times generate concise and idiomatic Chinese, and some segments were as good as strong human renditions. The central weakness was instability: locally successful output coexisted with sudden breakdowns into awkward or uninterpretable Chinese.

In the extemporaneous interview, the target-language profile changed more revealingly. Unintelligible or nonsensical output remained salient, but disfluency carryover increased sharply in most systems. This suggests that spontaneous discourse exposes a second target-language weakness that is much less visible in planned speech. When the source contains repetitions, hesitations, fillers, repairs, and loosely organised syntax, current MI systems more often reproduce these features too literally, resulting in heavier, more fragmented and harder-to-follow Chinese output. Human interpreters, by contrast, are more likely to regulate the surface form for the audience by compressing, smoothing, and reorganizing the discourse as they interpret. On this point, the present findings align with earlier work emphasizing the human advantage in audience-oriented communicative mediation (Liu & Liang, 2024).

A notable exception is System D. Unlike the other systems, it did not show the same rise in disfluency carryover in the unplanned text, suggesting that not literally reproducing all repetitions, fillers, and false starts may already be technologically attainable, for instance through automatic disfluency suppression (Fitzgerald et al., 2009). At the same time, System D did not thereby resolve the broader target-language-quality problem, because unintelligible or nonsensical output still remained substantial. The implication is more limited but useful: some aspects of spoken-disfluency control may be within reach, even though stable target-language quality in spontaneous discourse remains unsolved.

A final methodological point concerns the status of unintelligible or nonsensical output. In the present study, this code is retained under target-language quality even though it often also affects meaning, because unintelligibility is not only a failure of content transfer but directly a failure of target-language acceptability and usability. In real-time interpreting, a fragmented or nonsensical segment immediately increases the audience's processing burden, regardless of whether part of the intended meaning can still be inferred.

### 3.3.3 Illustrative Examples of Target-Language-Quality Problems

Three brief examples illustrate recurrent target-language weaknesses.

**Example 1. Unintelligible or nonsensical output**

Source: "To understand this, consider how Google's Sycamore quantum computer can solve a maths problem in 200 seconds that would take the world's fastest supercomputer 10,000 years."

System D output: "为了理解这一点，想想谷歌是如何成为参天大树的。考虑接下来的 100 秒。世界已经旋转了一万年。" [Back-translation: "To understand this, think about how Google became a towering tree. Consider the next 100 seconds. The world has been spinning for 10,000 years."]

The proper noun *Sycamore* is misrendered as 参天大树, the numerical information is distorted, and the final clause is semantically unrelated to the intended meaning. The output no longer conveys an interpretable proposition in Chinese.

**Example 2. Awkwardness/register**

Source: "So how…That's it's so interesting when you say that."

System C output: "所以当你这么说时，这真的很有意思。" / System E output: "所以你这么说的时候真是太有意思了。"

Both outputs follow the English surface structure too closely. A human interpreter would smooth this into a more conversational formulation, such as 你这么一说就很有意思了.

**Example 3. Disfluency carryover**

Source: “You know, I think, I think, you know, we probably, probably agreed with the vast, vast majority of it.”

System E output: “你知道，我认为，我觉得，你知道，我们可能大概率达成了广泛的共识。大部分内容。”

The system carries over multiple source disfluencies instead of filtering them. A more natural rendition would be 其中大部分内容, 可能我们都认同.

### 3.4 Discussion

#### 3.4.1 Boundary Conditions and Market Relevance

The present findings support a more differentiated reading of MI than either simple replacement narratives or blanket dismissals allow. The competitiveness of the strongest MI systems on planned speech suggests that under read-out, topically focused conditions delivered in standard accents, top systems may achieve meaning transfer competitive with portions of the professional human range.

This should be read as a bounded finding: the planned discourse materials were produced by native speakers with standard accents, establishing an upper-bound estimate of MI performance (see §3.5). This matters in practice, since real conferences rarely guarantee such controlled input. The results therefore point less to uniform replacement than to market segmentation: MI appears far more usable in structurally favourable discourse conditions, while human interpreters retain a clear advantage in spontaneous speech (§3.1). This is consistent with Fantinuoli’s (2025) expectation that, in the short term, outcome-level parity between machines and competent human interpreters “seems plausible only in constrained conditions: specific language pairs, well-defined settings, and predictable modalities” (p. 2). This discourse-conditioned picture contrasts with studies that aggregate across conditions and report a single performance gap.

A second qualification concerns what a competitive score means for users. The MEEC measures how much meaning is lost, not what kind of loss it is. On Text 2, System B and the strongest human interpreters reached similar scores through different errors: the machine distorted more, and the humans omitted more (§3.2.1). These errors may not carry the same risk for a listener. An omitted low-priority detail leaves the listener with less information. A distortion can leave the listener with wrong information, and nothing in the output signals that it is wrong. This risk may be greatest where MI is most competitive, because the stronger systems could at times produce concise and idiomatic Chinese on planned speech (§3.3). The present study did not test comprehension with real listeners, so this difference in risk is an inference from the error profiles, not a tested result. It nonetheless suggests that parity in aggregate score should not be read as parity in communicative reliability, and that comparisons of human and machine interpreting should report error profiles alongside total scores.

A further qualification concerns the scope of comparison itself. Even where machines become competitive on meaning, meaning transfer is not always the only relevant function of interpreting. Depending on the setting, the task may also involve institutional representation, audience design, and the adjustment of tone and register to the situation, all of which clients and institutions value beyond informational transfer alone. Fantinuoli (2025) similarly argues that, even if MI reaches functional parity, trust, accountability, and symbolic value may still favour human interpreters in particular settings.

### 3.4.2 Discussion for Interpreter Education

For interpreter education, the central implication lies in the discourse-mode contrast. If machines are catching up on the planned-speech side but remain substantially weaker in spontaneous discourse, training programmes that have historically concentrated on prepared, read-out speech may need rebalancing. Survey data from the 14 professional interpreters in the present study point in the same direction: 13 of the 14 reported that their typical assignments were either roughly balanced or weighted toward unscripted discourse (see Supplementary Methods, Daily-Assignment Discourse Mix Survey Item). This is a small sample and was not selected to represent the profession. The mix of scripted and unscripted speech may differ in other settings and markets. The urgency is underscored by the finding that, in the interview conditions, some advanced students still performed below the strongest MI systems.

Within the human cohort, all five interpreters who performed in the top quartile in both modes (H01, H05, H27, H28, H33) are professionals; no student in the present sample achieved this profile (Fisher's exact test, two-sided $p = .04$; see Supplementary Tables S9 and S10 for per-interpreter classification and contingency). Strong performance in both modes may therefore be acquired through training and experience, although a comparison of two cohorts cannot separate experience from selection into the profession. This adds to Liu et al.'s (2004) finding that interpreting expertise consists of domain-specific skill: in the present sample, that skill was uneven across discourse modes, and balance across the two was found only among professionals.

The cohort difference in cross-mode balance is paralleled by a tier-level difference in the structural pattern of code-level errors. Stronger human performers (top 8 by MEEC for each text) concentrated their losses in lower-priority minor-omission categories while preserving key-message content: top-tier means were 1.5 major omissions and 10.4 minor omissions (minor omission and minor list omission combined) per output on Text 2, and 1.3 major / 4.9 minor on Text 3. Weaker performers (bottom 8 by MEEC) showed many more major omissions in addition to minor ones (bottom-tier means: 9.0 major / 19.3 minor on Text 2; 5.0 major / 9.8 minor on Text 3; see Supplementary Table S12). The priority-graded distribution that distinguishes strategic from uncontrolled loss is therefore a property of stronger human performers, not of human interpreters as a category. This is consistent with the all-professional composition of the top-quartile-in-both cluster and with Liu et al.'s (2004) characterisation of expert interpreting as domain-specific selectivity skill.

The code-level findings specify which competences deserve priority. In spontaneous material, the machine disadvantage appeared primarily as meaning distortion, weak discourse linkage, disfluency carryover, and unstable target-language acceptability. These weaknesses correspond to the competences in which the strongest human interpreters most clearly retain their advantage: discourse tracking, contextual inference, speaker-intention reconstruction, and communicative mediation under pressure. Interpreter training programmes should accordingly identify each student's mode-specific weaknesses and target them explicitly, particularly in the unplanned mode where human interpreters retain their clearest advantage over MI.

At the same time, the code-level analysis revealed a systematic human weakness: under cognitive pressure, human interpreters were consistently more prone to omissions, particularly of list items, parallel structures, and lower-priority details. This omission tendency was observed across both discourse conditions. If appropriately integrated into the interpreting workflow, assistive technologies such as real-time ASR displays may have the potential to mitigate precisely this type of information loss, by offloading part of the monitoring burden and supporting more complete coverage of structured content (Ding et al., under review). A more realistic curricular direction is therefore dual: strengthen human performance where current MI remains weak, while also teaching students to use interpreting technologies strategically to compensate for documented human limitations under cognitive load (cf. Fantinuoli, 2018).

### 3.4.3 Discussion for MI/Speech Translation Development

For MI development, the results suggest that improvement paths differ by discourse condition. In planned speech, the post-hoc second-round attribution of machine major errors to ASR failure (§2.5.5) showed that a substantial proportion of residual major errors remained attributable to ASR-related problems. Further gains under favourable conditions may therefore depend primarily on ASR robustness: punctuation, boundary detection, and information-unit segmentation.

The spontaneous interview exposed a broader challenge. For the stronger systems, limitations could no longer be attributed to recognition failure alone; the deeper difficulty involved context tracking, turn parsing, disambiguation, and handling disfluent source speech. The target-language-quality data reinforce this conclusion: most systems showed sharply increased disfluency carryover in unplanned texts, although the System D pattern suggests some filtering may already be attainable. For spontaneous speech, the results suggest that progress will depend not only on more accurate speech recognition, but also on stronger context modelling and more selective decisions about what in the source speech carries communicative value. Among these gaps, speaker diarization (identifying which segments of audio correspond to different speakers) is a concrete and tractable target. None of the five systems tested in this study produced simultaneous-interpretation outputs that reflected speaker distinctions in the materials we tested. Example 2 above illustrates the consequence: when consecutive turns from different speakers are run together into a single utterance, the speaker boundary is lost in the translation, and the meaning may invert. Adding speaker-diarization support to consumer MI products would directly address this specific class of error. Ideally it would be a user-selectable option, since interactive and monologic input modes call for different parsing strategies.

### 3.5 Limitations and Future Research

The most important limitation concerns the source design. Each discourse mode is represented by one recording from one speaker, and the two texts in each mode are consecutive segments of the same recording. They are not independent replications of that mode. Although both recordings address the same broad topic, planning is bundled in these materials with speaker, speech rate, interactional structure and recording characteristics, and the design cannot separate them. The results show a marked contrast between this prepared speech and this interview, which is consistent with discourse sensitivity. Establishing a general discourse-mode effect would require additional speakers and recordings in each mode.

Several further limitations should be acknowledged. First, the analysis is restricted to English-to-Chinese simultaneous interpreting and should not be generalised uncritically to other language pairs. Second, the study did not evaluate delivery or latency as separate dimensions, focusing instead on output quality in terms of meaning and target-language acceptability. For MI systems that currently produce text-based output rather than synthesised speech, audience reception and the cognitive demands of reading real-time subtitles in interpreting contexts remain open questions for future research. Third, the dataset remains limited in scope and topic range. Finally, MI systems evolve rapidly; the results reported here represent a performance snapshot based on outputs generated on February 26–27, 2026, and subsequent updates to any system tested may shift its standing.

Another limitation concerns input conditions. The source materials were produced by native English speakers with standard accents, establishing a relatively favourable condition for MI and yielding an upper-bound estimate of system performance. Under more variable input conditions, such as non-native accents, regional variation, mixed-accent panels, or code-switching, MI performance would likely diverge further from the results reported here. The planned–unplanned contrast should also be understood as representing two points on a continuum rather than a strict dichotomy; speakers vary in delivery clarity regardless of preparation level, and the present sources (a polished scripted address and a relatively articulate but extemporaneous interview) may understate the variability within each mode.

A further methodological limitation concerns rater blinding. Although transcripts were anonymised and intermixed, complete blinding of output origin was unlikely, as discussed in §2.5.5. Future research should explore more controlled blinding protocols, such as presenting raters with only short excerpts rather than full transcripts or employing raters without prior exposure to MI output.

Distinct from rater blinding, a related limitation concerns rater configuration. The principal investigator served as one of two raters, and both raters had professional acquaintance with a subset of the 26 participants. This is a common constraint in conference interpreting research, where qualified raters often come from the same professional community as the participants. Procedural blinding (coded-file scoring; no consultation of the code-to-identifier mapping) and the high point estimates of pre-consensus inter-rater agreement (§2.5.2) mitigate but do not eliminate the risk of incidental voice recognition. The reliability check was also small. It used five outputs and two raters, fewer than the 30 samples and three raters that Koo and Li (2016) recommend where possible, so its confidence intervals are wide. This is partly offset because all reported scores come from consensus coding by both raters rather than from either rater alone. Future replications would benefit from rater pools sourced from outside the participants' immediate professional and pedagogical networks, and from a larger reliability sample.

These limitations point directly to a future research agenda. Particularly important are studies expanding the range of input conditions, testing a wider range of genres, and combining product-level evaluation with component-level analysis so that recognition failure, translation failure, and discourse-modelling limitations can be distinguished more precisely. Such research is needed to determine how robust the present study's central finding (the strong discourse sensitivity of MI relative to human interpreting) remains when it is tested outside controlled, standard-accent speech.

**4 Conclusion**

The central finding of this study is that the relative standing of MI systems changed sharply between a prepared speech and an extemporaneous interview. Under planned, read-out speech with standard accents, the strongest MI systems achieved meaning-transfer scores that overlapped with the professional human range on both planned texts. Under spontaneous, interactional speech, that overlap narrowed: professionals were markedly superior on both unplanned texts. On target-language quality, professional interpreters were ahead in both modes, and the gap widened on the interview. The best system outscored 11 of the 26 interpreters on each planned text and none on the interview, where most systems also carried over more of the speaker's disfluencies. This pattern is consistent with discourse mode acting as a boundary condition on MI competitiveness, although a design with one recording per mode cannot separate mode from other properties of the two sources (§3.5).

The code-level analyses of meaning and target language reveal why. Comparable aggregate scores in planned speech masked different error profiles: machines were distortion-prone where humans were omission-prone. For the listener, these errors differ in kind as well as in number: an omission withholds information, whereas a distortion supplies wrong information without warning. The specific character of machine error (ASR-linked misrecognition, context-insensitive lexical selection, weak discourse linkage) differed qualitatively from the strategic omissions and restructuring that characterised stronger human output. In spontaneous speech, these machine-specific vulnerabilities compounded, producing not merely more errors but a different kind of breakdown: failures of discourse tracking, contextual inference, and communicative coherence that current MI systems struggle to manage. What distinguished the stronger human interpreters under pressure was not the absence of error but where their errors fell: they lost lower-priority detail and kept the main propositions (cf. §3.4.2 and Supplementary Table S12). This pattern was evident among the top human performers but not uniformly across the cohort.

These findings carry implications in three directions. For the interpreting market, they suggest not uniform displacement but discourse-conditioned segmentation: MI may be viable for structured, predictable speech events, while human interpreters retain a clear advantage in spontaneous, interactional, and context-dependent communication. Users therefore need calibrated expectations: the same system that performs competitively during a scripted keynote may produce misleading output during an unscripted interview, Q&A session, or panel discussion. For interpreter education, the priority shifts toward the competences in which stronger human interpreters hold the greatest advantage: discourse tracking, speaker-intention inference, and disfluency management. For MI development, the results point to a dual bottleneck: ASR robustness in planned speech, and context modelling and discourse-level meaning construction in spontaneous speech. Whether the boundary conditions identified in this English–Chinese study generalise across typologically different language pairs, accent-variable input, and live multi-speaker settings remains the most pressing question for future research.

**Use of AI Tools**

In preparing this manuscript, the author used generative AI tools, Anthropic's Claude and OpenAI's GPT, for language polishing and copyediting, summarizing prior literature, compiling and formatting the reference list from author-provided source PDFs, and formatting the manuscript. Claude was also used to revise the R scripts included in the data deposit and, in November 2026, to write the R scripts for the two sensitivity analyses (Supplementary Tables S13 and S14) and for the target-language position counts in §3.3.1. The author reviewed and verified all AI-assisted output and takes full responsibility for the content of this article.

**Acknowledgements**

This study was supported by faculty development funding from the Middlebury Institute of International Studies at Monterey (MIIS). The author thanks Yuchen Yan and Leo Dai for their assistance with the machine-interpreting experiments, the collection of screen recordings of machine-interpreting output, the preparation of transcripts of machine-interpreting and human-interpreting output, and the literature search. The author also thanks Rachel Halliday from the META Lab at MIIS for her assistance with quantitative data analysis. Finally, the author thanks the professional interpreters and graduate interpreting students who took part in this study for their time and generosity.

**Author Contributions**

Yuanyuan Ding is the sole author and was responsible for all aspects of the study, including conception and design, development of the analytical framework, assessment, data analysis, and writing of the manuscript.

**Competing Interests**

The author declares no competing interests.

**Data Availability**

The de-identified text materials, scoring data, coded errors, instruments, and analysis scripts supporting the findings of this study will be made openly available in the Harvard Dataverse repository upon publication of the article. Until then, they are available from the author on reasonable request. The deposit will contain the four English source-language transcripts, the text outputs of the five machine-interpreting systems, the 20 partial human-interpreting transcripts produced for the machine-blinding round (§2.5.5), aggregate MEEC and target-language quality scores per output × text, per-text counts of major and minor errors, category-level meaning codes for every human and machine output on all four texts, the error taxonomy and rating rubric (Supplementary Appendices A and B), participant-variables documentation, and the R analysis scripts used to compute the statistics and figures reported here. A complete file inventory is provided in the deposit's README.

Original audio recordings of the human interpreting performances are not publicly deposited because human voice constitutes a biometric identifier, and the IRB-approved informed consent authorised use and sharing of de-identified data but did not explicitly authorise public release of voice recordings. De-identified audio is available to qualified researchers on reasonable request, subject to a data-sharing agreement and recipient-IRB review.

**Ethics Declaration**

The interpreting dataset analysed in this study was collected under a parent protocol, "Simultaneous Interpreting and Automatic Speech Recognition" (Protocol #560), reviewed by the Middlebury College Institutional Review Board (Middlebury, Vermont, USA) and granted exempt status on 28 January 2026 under 45 CFR 46.101(b)(2)–(3). All research procedures were performed in accordance with the U.S. Common Rule (45 CFR 46) and the World Medical Association's Declaration of Helsinki. The present manuscript reports a secondary analysis of de-identified human-interpreting outputs originally collected under this approved protocol; no additional data were collected.

Written informed consent was obtained electronically from all adult participants prior to data collection, using an IRB-approved consent form. The consent procedure disclosed the study purpose, voluntary participation, the right to withdraw without penalty, and the use of participant codes for de-identification. Participants additionally consented to the retention and future research use of their de-identified data, which authorises the secondary analysis reported here.

---

[i] Input conditions were comparable across human and machine conditions: human interpreters received source audio via Zoom (system audio feed), while MI systems received the same recordings routed through system audio. The ASR tool used in the human ASR-assisted condition (Microsoft ASR) is distinct from the proprietary ASR modules built into each MI system; variation in built-in ASR quality is treated as part of each system's overall performance profile rather than as a confound, because this study evaluates MI systems as integrated, user-facing products.

# Supplementary Information

## Machine interpreting competitiveness: a discourse-sensitive comparison of meaning and target-language quality in English–Chinese simultaneous interpreting

Yuanyuan Ding

Associate Professor, Middlebury Institute of International Studies at Monterey

Correspondence: yuanyuand@middlebury.edu; cherrydyy@gmail.com

## Supplementary Methods

### Machine Interpreting System Selection: Excluded Systems

Several other prominent speech translation technologies were considered but did not meet the inclusion criteria. Google's real-time translation extension generates only monolingual target-language captions without parallel source transcripts, precluding the bilingual output comparison required by the present evaluation design. Google Meet has since introduced a speech translation feature, but at the time of data collection it remained in beta, available only to certain paid subscribers, and subject to significant platform limitations: translation is available only on desktop, not on mobile devices; participants joining through Meet hardware can listen to translations but cannot have their speech translated; and the feature is not available in live streams or recordings. Meta's SeamlessM4T and OpenAI's real-time speech API are designed as developer toolkits requiring programmatic implementation rather than functioning as ready-to-use interpreting products accessible to the general public. Additionally, platforms such as Wordly and KUDO were considered but excluded because they operate as enterprise-contracted services available only through institutional licensing and sales-mediated onboarding rather than as publicly accessible consumer products; at the time of data collection, neither offered a self-service product that an individual user could independently access and evaluate under standardised conditions. It is worth noting that an independent evaluation conducted by the WHO interpretation team found Wordly's interpretation into Chinese to be well below acceptable quality thresholds in a multilingual institutional setting (WHO 2025). Because this study aims to evaluate MI systems as they are encountered by end users and professional interpreting clients in real-world settings, the inclusion criteria were restricted to products that provide an integrated, publicly accessible, user-facing simultaneous interpreting experience with extractable bilingual output.

**Tested MI System Specifications**

To support replicability and specify the precise versions and configurations of the five commercial machine-interpreting systems whose outputs are reported in this study, we document below the platform, version, and user-selectable settings of each system as accessed during data collection in February 2026. All systems were accessed in their publicly available, user-facing forms, consistent with the selection criterion described in §2.2 of the main manuscript; no developer APIs, enterprise-contracted services, or domain-tuned deployments were used.

Tencent Meeting was accessed through the desktop application (v3.41.1). The transcription feature was enabled. Source language was set to automatic recognition; translation language to bidirectional English–Chinese; display mode to show both source and translation; industry domain to general.

Microsoft Teams Premium was accessed through the web application under a 60-day Premium trial subscription. The live captioning feature was enabled. Transcription source language was set to English; translation language to Chinese; display mode to show both source and translation in real-time.

Doubao Simultaneous Interpretation 2.0 was accessed through the Volcengine Ark web-based experience platform, under the model identifier doubao-simul-s2s. Speech-to-speech (S2S) mode was enabled with the simultaneous interpretation (SI) tab active. Source language was set to English; target language to Chinese.

iFlyrec (Spark Translation) was accessed through the web application, in its Personal Edition (个人版), via the cloud-based interface. The Spark simultaneous interpretation feature was enabled. Audio input was set to system/microphone; source language to English; target language to Chinese; the smart correction setting was active.

Youdao (NetEase) was accessed through the mobile application (v4.3.37). The AI Simultaneous Interpretation feature was enabled. Target language was set to Chinese; the adaptive noise reduction setting was active.

Where settings of analogous function existed across systems (for example, source-/target-language designation, display mode, audio-input source), they were configured to provide English-to-Chinese SI with on-screen output where applicable. System-specific options without analogues across all five systems (for example, Tencent's industry: general domain selector, iFlyrec's smart correction, Youdao's adaptive noise reduction) were configured to default or general-purpose settings rather than to domain- or task-tuned options, in order to maintain comparability with what an ordinary user would receive.

**MI System Functional Configurations**

Among the five systems examined, only iFlyrec and Doubao provide integrated speech synthesis for target-language output, whereas Microsoft Teams, Youdao, and Tencent primarily operate in a speech-to-text mode, generating written translations without audio output.

For the purposes of the present study, evaluation focuses exclusively on meaning and target-language quality. The delivery dimension (e.g., prosody, fluency, and voice quality) was not included, as comparable audio output is not consistently available across systems.

Crucially, all five systems are public-facing products designed for end users, not developer-oriented tools. Microsoft Teams and Tencent integrate simultaneous interpreting functionality directly into their meeting platforms, while iFlyrec, Youdao, and Doubao are marketed explicitly as simultaneous interpreting products.

**Human Interpreter Participation and Pooling Rationale**

All interpreters participated via Zoom under identical conditions: source audio was delivered through Zoom, and interpreters recorded their renditions on a dedicated recording platform. Participants without a quiet space at home could join from an interpreting booth on campus; the booth served only as a quiet room, and no booth-installed equipment was used for audio input or output.

Each participant interpreted two texts with ASR assistance and two without, one of each per discourse mode. On each text, half the outputs in each cohort were ASR-assisted (professionals: 7 with ASR, 7 without; students: 6 with ASR, 6 without). For the main analysis, ASR-assisted and unassisted performances are combined within each cohort, for three reasons:

1. Tool-assisted interpreting remains human interpreting. The interpreter retains full agency, responsibility, and decision-making authority regardless of whether ASR captions are available. Unlike MI, where the system autonomously generates the target-language output, a human interpreter using ASR is still responsible for comprehension, reformulation, and delivery.
2. ASR support is a growing part of interpreting technology. Fantinuoli (2018) places it within a broader "technological turn" in interpreting, in which computer-assisted interpreting tools can draw on automatic speech recognition; pooling thus captures the realistic range of contemporary human performance against which MI systems should be evaluated.

3. The main pattern does not depend on pooling. Each MI system was compared separately with the unassisted and the ASR-assisted human outputs on each text (13 outputs per subset; Supplementary Table S13). On both planned texts, System B and System E outperformed most human outputs in both subsets. On both unplanned texts, they outperformed fewer than half. Pooling did not shift the benchmark in a single direction: on Texts 1 to 3, the three competitive systems outperformed fewer assisted than unassisted outputs, and on Text 4 the reverse held. The subsets are small, so these counts are descriptive.

Each output's ASR condition is recorded in the data deposit.

**Supplementary Table S13. Number of human outputs outperformed by each MI system, shown separately for unassisted and ASR-assisted human outputs.**

**Panel A. All human outputs (n = 13 per subset)**

| System | Human subset | Text 1 | Text 2 | Text 3 | Text 4 |
|---|---|---|---|---|---|
| System B | Unassisted | 13 | 11 | 3 | 1 |
| System B | ASR-assisted | 9 | 10 | 0 | 5 |
| System E | Unassisted | 10 | 9 | 2 | 4 |
| System E | ASR-assisted | 9 | 7 | 0 | 6 |
| System C | Unassisted | 10 | 7 | 2 | 1 |
| System C | ASR-assisted | 8 | 2 | 0 | 3 |
| System A | Unassisted | 0 | 0 | 0 | 0 |
| System A | ASR-assisted | 0 | 0 | 0 | 0 |
| System D | Unassisted | 0 | 0 | 0 | 0 |
| System D | ASR-assisted | 0 | 0 | 0 | 0 |

**Panel B. Professional interpreters only (n = 7 per subset)**

| **System** | **Human subset** | **Text 1** | **Text 2** | **Text 3** | **Text 4** |
|---|---|---|---|---|---|
| System B | Unassisted | 7 | 5 | 1 | 1 |
| System B | ASR-assisted | 3 | 4 | 0 | 0 |
| System E | Unassisted | 4 | 3 | 0 | 2 |
| System E | ASR-assisted | 3 | 3 | 0 | 0 |
| System C | Unassisted | 4 | 1 | 0 | 1 |
| System C | ASR-assisted | 2 | 1 | 0 | 0 |
| System A | Unassisted | 0 | 0 | 0 | 0 |
| System A | ASR-assisted | 0 | 0 | 0 | 0 |
| System D | Unassisted | 0 | 0 | 0 | 0 |
| System D | ASR-assisted | 0 | 0 | 0 | 0 |

Note. A human output counts as outperformed when its MEEC is higher than the system's. Texts 1 and 2 are planned discourse; Texts 3 and 4 are unplanned discourse. Reproducible from Supplementary_Data.xlsx and Meaning_Codes.csv using ASR_subset_sensitivity.R in the data deposit.

### Sensitivity to Error Weighting

The main analysis weights five minor errors as one major error (5:1). To check whether the results depend on this choice, each MI system's position among the 26 human outputs was recomputed with major errors only and with 3:1 and 10:1 weightings (Supplementary Table S14). Under every weighting, System B outperformed 15 to 22 human outputs on the planned texts and 3 to 6 on the unplanned texts. System E outperformed 9 to 12 human outputs on Text 4 across the tested weightings; the largest position changes occurred on Text 2.

**Supplementary Table S14. Number of the 26 human outputs outperformed by each MI system under alternative error weightings.**

| System | Weighting | Text 1 | Text 2 | Text 3 | Text 4 |
|---|---|---|---|---|---|
| System B | Major errors only | 21 | 15 | 3 | 6 |
| System B | 3:1 | 22 | 22 | 3 | 5 |
| System B | 5:1 (main analysis) | 22 | 21 | 3 | 6 |
| System B | 10:1 | 22 | 18 | 3 | 6 |
| System E | Major errors only | 18 | 10 | 1 | 12 |
| System E | 3:1 | 19 | 17 | 2 | 9 |
| System E | 5:1 (main analysis) | 19 | 16 | 2 | 10 |
| System E | 10:1 | 19 | 11 | 2 | 11 |
| System C | Major errors only | 18 | 6 | 3 | 6 |
| System C | 3:1 | 16 | 9 | 2 | 2 |
| System C | 5:1 (main analysis) | 18 | 9 | 2 | 4 |
| System C | 10:1 | 18 | 9 | 3 | 6 |
| System A | Major errors only | 0 | 0 | 0 | 2 |
| System A | 3:1 | 0 | 0 | 0 | 0 |
| System A | 5:1 (main analysis) | 0 | 0 | 0 | 0 |
| System A | 10:1 | 0 | 0 | 0 | 0 |
| System D | Major errors only | 0 | 0 | 0 | 0 |
| System D | 3:1 | 0 | 0 | 0 | 0 |
| System D | 5:1 (main analysis) | 0 | 0 | 0 | 0 |
| System D | 10:1 | 0 | 0 | 0 | 0 |

Note. Values are the number of the 26 human outputs whose score was higher than the system's. Texts 1 and 2 are planned discourse; Texts 3 and 4 are unplanned discourse. Reproducible from Error_Counts.xlsx using MEEC_weight_sensitivity.R in the data deposit.

**Daily-Assignment Discourse Mix Survey Item**

To characterise the discourse mix encountered in the professional cohort's typical professional work, the pre-task questionnaire (administered under Middlebury College IRB Protocol #560) included a single item asking professional interpreters about the composition of their source-speech assignments. Item Q47 read: "In your typical interpreting assignments, the source speech is mostly:" Responses were collected on a seven-point scale, ordered from most spontaneous to most scripted, with each band defined by an approximate proportion of unscripted versus scripted source speech.

**Supplementary Table S11. Distribution of professional-cohort responses to Item Q47 (daily-assignment discourse mix), n = 14.**

| **Response option** | **Count** |
|---|---|
| Almost always speaking spontaneously (90–100% unscripted) | 1 |
| Usually speaking spontaneously (70–89% unscripted) | 4 |
| Somewhat more unscripted/spontaneous (55–69% unscripted) | 4 |
| Roughly balanced (45–54% scripted) | 4 |
| Somewhat more prepared speeches/read aloud (55–69% scripted) | 1 |
| Usually reading prepared text (70–89% scripted) | 0 |
| Almost always reading prepared text (90–100% scripted) | 0 |
| Total | 14 |

Thirteen of 14 respondents (92.9%) reported typical assignments at the "roughly balanced" point or weighted more toward unscripted/spontaneous source speech (the upper four rows of the distribution); the remaining respondent reported assignments somewhat more weighted toward prepared/read-aloud source speech. The two most-scripted options received no responses. This distribution is consistent with the point made in §3.4.2 of the main manuscript: for these 14 respondents, unscripted speech made up a large share of typical assignments. The sample is small and was not selected to represent the profession.

**Source Material Descriptors**

**Readability indices.**

Readability indices (Flesch Reading Ease and Flesch–Kincaid Grade Level) are reported as conventional descriptors of lexical and syntactic complexity. Planned discourse: Text 1 (Flesch Reading Ease = 59.74; Flesch–Kincaid Grade Level = 9.76), Text 2 (Flesch Reading Ease = 60.92; Flesch–Kincaid Grade Level = 9.48). Unplanned discourse: Text 3 (Flesch Reading Ease

= 75.56; Flesch–Kincaid Grade Level = 6.63), Text 4 (Flesch Reading Ease = 77.91; Flesch–Kincaid Grade Level = 5.86). However, these measures were developed for assessing written text comprehension by readers and do not directly predict the difficulty of simultaneous interpreting, which depends on the interaction of multiple factors including delivery rate, informational density, and cognitive load (Gile 2009). The readability scores are therefore provided for descriptive comparability with prior studies, not as indicators of interpreting difficulty.

**Speech rates.**

After the 10% speed reduction applied in the parent study (Ding et al. under review), the planned source ran at approximately 140 WPM and the unplanned source at approximately 170 WPM. The planned rate lies just below the average of about 150 WPM documented at the UN Human Rights Council (Barghout et al. 2012, as cited in Seeber 2015), and the unplanned rate falls within the average range that Tauroza and Allison (1990) report for spoken interviews (160–210 WPM). Although the unplanned discourse segments exhibit a higher nominal rate, speech rate alone does not determine the difficulty of simultaneous interpreting. Drawing on Gile's (2009) Effort Models, processing difficulty arises from the interaction of delivery rate with informational density and cognitive load. The unplanned discourse segments' lower informational density and greater use of discourse markers may partly offset their higher rate.

**Ecological validity.**

The inclusion of two discourse modes, with two texts per mode, addresses a key limitation in prior research on interpreting technology, which has rarely treated discourse mode as a variable, thereby constraining ecological validity. Liu and Liang (2024), for example, drew all their source texts from speeches by Chinese government leaders at international conferences. More broadly, the selection reflects typical communicative demands in contemporary English–Chinese interpreting practice, where interpreters are routinely required to process both planned, pre-formulated discourse and spontaneous, unplanned speech within high-stakes institutional and commercial contexts.

## Rater Training Details

Rater training followed established best practices for rater-mediated assessment, including: calibration using benchmark samples; alignment on rubric descriptors; and consensus-building on error classification and severity. During consensus coding, one rater served as the primary coder while the second monitored in real time; when the second rater identified a disagreement, coding was paused and the two discussed the annotation until agreement was reached.

Inter-rater reliability was assessed on a subset of five human interpreting outputs using intraclass correlation coefficients (ICC; two-way mixed, absolute agreement, single measures). Point estimates of agreement were high for both dimensions: meaning accuracy (major-equivalent error counts), ICC(A,1) = .970, 95% CI [.741, .997], $p = .001$; target-language quality (100-point scale), ICC(A,1) = .926, 95% CI [.270, .992], $p = .001$ (see Supplementary Table S8 for full ICC output). Because only five outputs were rated, the confidence intervals are wide. Following Koo and Li (2016), who recommend judging reliability by the confidence interval rather than the point estimate, agreement ranged from moderate to excellent for meaning accuracy and from poor to excellent for target-language quality. Having established rater alignment during this pilot phase, the two raters then evaluated all remaining outputs using the dimension-specific procedures described in the main text.

**Blinding Procedures**

It should be acknowledged that complete blinding was unlikely to have been achieved. Some machine outputs—particularly from weaker-performing systems, and more so in the unplanned discourse condition, where disfluency carryover and unintelligible segments are distinctive—were probably identifiable to experienced raters. At the same time, the stronger MI systems on planned speech, which produced locally fluent and natural-sounding Chinese, were less obviously distinguishable from human output; blinding was therefore partially rather than wholly compromised. Fantinuoli and Prandi (2021) similarly describe blatant errors arising from speech recognition as “a distinctive characteristic of MI”.

Three features of the evaluation design mitigate the risk of systematic bias. First, all evaluations were conducted using a structured analytic rubric with pre-calibrated descriptors, constraining rater judgement to specific, observable features of meaning accuracy and target-language quality rather than global impressions of output origin. Second, machine and human transcripts were intermixed within each evaluation batch, so raters could not approach them as categorically different sets. Third, the direction of any residual bias is conservative for one finding but not for the other. If raters unconsciously applied stricter standards to suspected machine output, the MI systems’ competitive performance on planned speech would be understated rather than overstated. The planned–unplanned contrast is less protected: machine output was easier to recognise in unplanned discourse, so stricter scoring there could widen the contrast. The structured rubric and the code-level evidence limit this risk but cannot exclude it.

**Transcript-Based Evaluation Validity**

Because most machine interpreting (MI) systems generate textual output rather than audio, all evaluations of machine performance in the present study were conducted using standardised three-column transcripts, including: (1) the source English text (accuracy verified by expert reviewers), (2) machine-generated ASR output, and (3) the Chinese output (machine-generated translations or verbatim transcripts of human interpreting).

Input conditions were comparable across human and machine conditions. In the original study, human interpreters received the source audio via videoconferencing software (system audio feed, not external playback); MI systems likewise received the same source audio routed directly through system audio. Both conditions thus used the same recordings under equivalent digital audio delivery.

It should be noted that the ASR tool used in the human ASR-assisted condition (Microsoft ASR) is distinct from the built-in ASR modules employed by the five MI systems, each of which relies on its own proprietary speech recognition component as part of its processing pipeline. Variation in ASR quality across MI systems was indeed observed and is itself a relevant feature of the products under evaluation. Because this study evaluates MI systems as integrated, user-facing products rather than isolating individual pipeline components, differences in built-in ASR quality are treated as part of the system's overall performance profile rather than as a confound.

In the original ASR study, human performances were evaluated primarily based on audio, using two-column transcripts (source text and ASR output). To check whether evaluation format affected scores, the same 20 human performances (five per source text) were re-scored from three-column transcripts using the same criteria. Target-language quality scores were virtually unchanged. Individual meaning codes varied somewhat between the two passes, as is expected when the same performance is coded twice, but the resulting MEEC scores remained close. Because the MEEC rather than individual codes is the unit of analysis, the original audio-based human scores were retained.

**Supplementary Statistical Analysis**

The main text reports descriptive statistics and qualitative analysis, consistent with the exploratory design and sample sizes. The following non-parametric analyses are provided as a supplementary robustness check to contextualise the descriptive patterns reported in Section 3 of the main text. All tests are two-tailed; significance is noted at $p < .05$ (*) and $p < .01$ (**). No corrections for multiple comparisons are applied, and results should be interpreted accordingly.

**Supplementary Table S6. Descriptive statistics for MEEC by group and text**

| Group | n | Text 1 (planned) | | Text 2 (planned) | | Text 3 (unplanned) | | Text 4 (unplanned) | |
|---|---|---|---|---|---|---|---|---|---|
| | | ***M*** | ***Mdn*** | ***M*** | ***Mdn*** | ***M*** | ***Mdn*** | ***M*** | ***Mdn*** |
| Top 3 MI | 3 | 8.7 | 8.8 | 11.2 | 10.8 | 15.0 | 15.0 | 16.7 | 17.4 |
| All MI | 5 | 16.4 | 10.4 | 19.3 | 15.8 | 17.3 | 15.6 | 19.2 | 18.2 |
| Professional | 14 | 11.1 | 9.2 | 9.9 | 8.6 | 7.9 | 7.6 | 9.1 | 8.8 |
| Student | 12 | 19.3 | 19.7 | 16.6 | 19.2 | 11.8 | 11.2 | 15.7 | 16.7 |

*Note.* Top 3 MI = System B, System E, System C (the three strongest systems identified in Section 3.1 of the main text). MEEC = major-equivalent error count; lower values indicate better meaning accuracy.

**Between-group comparisons per text.** Mann–Whitney U tests were used to compare groups on each text individually (Supplementary Table S7). Consistent with the main text's descriptive findings, the top three MI systems did not differ significantly from professional interpreters on either planned-discourse text (Text 1: $U = 19.0$, $p = .850$; Text 2: $U = 25.0$, $p = .659$), but produced significantly more errors on both unplanned-discourse texts (Text 3: $U = 40.5$, $p = .016$; Text 4: $U = 38.0$, $p = .038$). The rank-biserial effect sizes confirm the pattern: near-zero on planned discourse (Text 1: $r = 0.10$; Text 2: $r = -0.19$) and large on unplanned discourse (Text 3: r = −0.93; Text 4: r = −0.81). Professional interpreters significantly outperformed advanced students on all four texts.

**Supplementary Table S7. Mann–Whitney U tests comparing groups by text**

| Comparison | Text 1 (planned) | | Text 2 (planned) | | Text 3 (unplanned) | | Text 4 (unplanned) | |
|---|---|---|---|---|---|---|---|---|
| | ***U*** | ***p*** | ***U*** | ***p*** | ***U*** | ***p*** | ***U*** | ***p*** |
| Top 3 MI vs. Prof | 19.0 | .850 | 25.0 | .659 | **40.5** | **.016*** | **38.0** | **.038*** |
| Prof vs. Student | **25.5** | **.003**** | **32.5** | **.009**** | **37.5** | **.018*** | **25.0** | **.003**** |
| All MI vs. All human | 71.0 | .768 | 84.0 | .320 | **122.5** | **.002**** | **110.0** | **.017*** |

*Note.* Top 3 MI = System B, System E, System C. All MI includes all five systems. Bold indicates significance. *p < .05. **p < .01.

**Positioning of individual MI systems.** On planned discourse, the best-performing system (System B, MEEC = 7.0) outperformed 57% of professional interpreters, and the second-best (System E, MEEC = 9.8) outperformed 50%. On unplanned discourse, no MI system outperformed more than 7% of professionals. This shift from competitive positioning to near-complete separation further underscores the discourse-conditioned nature of MI competitiveness.

**Caveats.** These analyses should be interpreted with caution. The MI sample (n = 5; n = 3 for top systems) represents specific commercial products rather than a random sample from a population of systems; the between-group tests therefore describe the observed data rather than licensing broad generalisations about machine interpreting as a category. The within-group comparisons for human cohorts are better powered (n = 14 and n = 12) but remain limited in statistical power. All reported p-values are uncorrected for multiple comparisons.

**Supplementary Table S8. Intraclass correlation coefficients for inter-rater reliability**

Two-way mixed-effects, absolute-agreement, single-measure ICC. Assessed on a subset of five human interpreting outputs.

***Panel A. Meaning accuracy (major-equivalent error counts)***

| **Measure** | **ICC** | **95% CI lower** | **95% CI upper** | ***F*** | ***$df_1$*** | ***$df_2$*** | ***p*** |
|---|---|---|---|---|---|---|---|
| Single measures | .970 | .741 | .997 | 53.4 | 4 | 4 | .001 |
| Average measures | .985 | .851 | .998 | 53.4 | 4 | 4 | .001 |

***Panel B. Target-language quality (100-point scale)***

| **Measure** | **ICC** | **95% CI lower** | **95% CI upper** | ***F*** | ***$df_1$*** | ***$df_2$*** | ***p*** |
|---|---|---|---|---|---|---|---|
| Single measures | .926 | .270 | .992 | 51.0 | 4 | 4 | .001 |
| Average measures | .962 | .412 | .996 | 51.0 | 4 | 4 | .001 |

*Note.* ICC = intraclass correlation coefficient; CI = confidence interval. Two-way mixed-effects, absolute-agreement model (Koo and Li 2016). Single measures correspond to ICC(A,1); average measures correspond to ICC(A,k) with k = 2 raters.

## Cohort × cross-mode performance pattern

Within the human cohort (n = 26), interpreters were classified by their performance in both planned (mean of Texts 1–2) and unplanned (mean of Texts 3–4) discourse modes. Three groups were defined by quartile membership: Group 1 (top quartile in both modes), Group 2 (mixed/middle), and Group 3 (bottom quartile in both modes). The 'top quartile in both' configuration is concentrated in the professional cohort (Supplementary Tables S9 and S10): all five interpreters in Group 1 are professionals; no student in the present sample achieved this profile. Fisher's exact test on cohort × Group-1 membership returned a two-sided p = .04. The configuration was thus associated with professional status in this sample; the design cannot separate training and experience from selection into the profession (see §3.4.2 of the main manuscript).

**Supplementary Table S9. Per-interpreter classification of all 26 human interpreters by cross-mode performance pattern.**

| Output_ID | Cohort | Planned mean | Unplanned mean | Group |
|---|---|---|---|---|
| H33 | Professional | 5.20 | 3.40 | 1: Top-quartile in both |
| H05 | Professional | 6.30 | 2.30 | 1: Top-quartile in both |
| H27 | Professional | 5.20 | 4.10 | 1: Top-quartile in both |
| H01 | Professional | 6.80 | 6.50 | 1: Top-quartile in both |
| H28 | Professional | 9.10 | 6.20 | 1: Top-quartile in both |
| H19 | Professional | 6.20 | 9.50 | 2: Mixed/middle |
| H35 | Professional | 6.10 | 10.80 | 2: Mixed/middle |
| H26 | Professional | 11.30 | 5.70 | 2: Mixed/middle |
| H30 | Professional | 10.60 | 8.60 | 2: Mixed/middle |
| H15 | Professional | 12.80 | 9.00 | 2: Mixed/middle |
| H23 | Professional | 11.30 | 11.00 | 2: Mixed/middle |
| H20 | Student | 12.30 | 10.20 | 2: Mixed/middle |
| H09 | Student | 14.50 | 11.30 | 2: Mixed/middle |
| H08 | Professional | 14.90 | 11.00 | 2: Mixed/middle |
| H18 | Student | 14.80 | 12.10 | 2: Mixed/middle |
| H29 | Student | 15.90 | 11.10 | 2: Mixed/middle |
| H07 | Student | 16.90 | 10.20 | 2: Mixed/middle |
| H02 | Student | 15.50 | 16.60 | 2: Mixed/middle |
| H17 | Student | 18.90 | 14.60 | 2: Mixed/middle |
| H37 | Professional | 20.00 | 14.40 | 2: Mixed/middle |
| H24 | Student | 19.50 | 15.00 | 3: Bottom-quartile in both |
| H12 | Student | 21.20 | 15.30 | 3: Bottom-quartile in both |
| H13 | Student | 23.00 | 14.20 | 2: Mixed/middle |
| H03 | Student | 22.90 | 15.60 | 3: Bottom-quartile in both |
| H10 | Professional | 21.50 | 17.10 | 3: Bottom-quartile in both |
| H06 | Student | 20.00 | 18.90 | 3: Bottom-quartile in both |

Note. Group 1 (top-quartile in both modes) defined as planned mean ≤ 9.47 and unplanned mean ≤ 8.70. Group 3 (bottom-quartile in both modes) defined as planned mean > 19.35 and unplanned

mean > 14.55. Group 2 contains all interpreters not in Groups 1 or 3. Rows are sorted by combined (planned + unplanned) MEEC, lowest to highest.

**Supplementary Table S10. Cohort × Group 1 (top-quartile in both modes) contingency, with Fisher's exact test result.**

| Cohort | In Group 1 (top-quartile in both) | Not in Group 1 | Total |
|---|---|---|---|
| Professional | 5 | 9 | 14 |
| Student | 0 | 12 | 12 |
| Total | 5 | 21 | 26 |

Note. The 5 vs. 0 split in the top-quartile-in-both category is statistically significant by Fisher's exact test (two-sided $p = .04$; one-sided, professionals > students, $p = .03$). The Group 3 (bottom-quartile-in-both) split (1 of 14 professionals vs. 4 of 12 students) is in the predicted direction but does not reach significance with the present sample size (Fisher's exact, two-sided $p = .15$). Reproducible via the deposit script MEEC_inferential_analysis.R.

**Robustness to within-text scaling.** Because MEEC counts depend in principle on per-text error opportunity (length, density, propositional load), the Group 1 classifier was re-run using a scale-free alternative: each interpreter's rank within the human pool ($n = 26$) on each text, averaged within mode (planned rank = mean of ranks on Texts 1 and 2; unplanned rank = mean of ranks on Texts 3 and 4). The same five interpreters (H01, H05, H27, H28, H33; all professionals) and zero students are identified as Group 1 under both classifiers, so the 2 × 2 contingency table and Fisher's exact test are numerically identical (Haldane–Anscombe corrected OR = 14.5; two-sided $p = .0425$; one-sided $p = .0304$). The substantive conclusion that balanced cross-mode strength is observed exclusively among professionals is therefore robust to whether mean MEEC values or within-text mean ranks are used to define top-quartile membership.

**Supplementary Table S12. Major versus minor omission rates by human performance tier, Texts 2 and 3.**

The cohort × cross-mode performance pattern is paralleled by a tier-level structural pattern in human-side errors. Among the 26 humans, performance tier (defined by within-text MEEC rank) is associated not only with overall MEEC but with the distribution of errors between major-meaning categories (key-message content lost) and minor categories (list items, parallel structures, supporting details). The major-omission rate differentiates strategic from uncontrolled loss directly: top-tier humans concentrate their losses in minor categories while preserving the main propositions, whereas bottom-tier humans show many more major omissions alongside minor ones.

**Panel A. Text 2 (planned discourse)**

| **Performance tier** | **n** | **Mean MEEC** | **Mean major omissions** | **Mean minor omissions (total)** |
|---|---|---|---|---|
| Top tier (rank 1–8) | 8 | 6.0 | 1.5 | 10.4 |
| Mid tier (rank 9–18) | 10 | 12.1 | 5.0 | 16.4 |
| Bottom tier (rank 19–26) | 8 | 21.2 | 9.0 | 19.3 |

**Panel B. Text 3 (unplanned discourse)**

| Performance tier | n | Mean MEEC | Mean major omissions | Mean minor omissions (total) |
|---|---|---|---|---|
| Top tier (rank 1–8) | 8 | 5.2 | 1.3 | 4.9 |
| Mid tier (rank 9–18) | 10 | 9.7 | 3.0 | 7.0 |
| Bottom tier (rank 19–26) | 8 | 14.3 | 5.0 | 9.8 |

Note. Top tier = top 8 human interpreters by MEEC for each text; mid tier = ranks 9–18; bottom tier = ranks 19–26. Mean major omissions and mean minor omissions are per-output averages within each tier. The 4–6× difference in major-omission rate across tiers, against a much smaller difference in minor-omission rate, is the empirical basis for the discussion-section claim (§3.4.2 of the main manuscript) that priority-graded loss is concentrated among stronger human performers. Reproducible from the category-level code counts in Meaning_Codes.csv (deposited in Harvard Dataverse).

**Supplementary Note: Additional Illustrative Examples of Machine Error Patterns**

The following examples supplement the four illustrative cases presented in Section 3.2.3 of the main text. They are drawn from the planned and unplanned discourse texts and exemplify additional recurrent patterns by which MI output deteriorates.

**Clause boundary error leading to propositional fragmentation**

Source: "That's why the UK is so well-placed to create the world's first Safety Institute."

System E ASR: "That's why the UK is so well placed to create. The world's first AI safety institute."

System E output: "这就是为什么英国在创新方面具有得天独厚的优势。全球首个人工智能安全研究所。"

**Discussion:**

The ASR system inserts a sentence boundary after create, breaking a single proposition into two fragments. The resulting Chinese output no longer expresses the intended causal relation that the UK is well placed to create the institute; instead, it yields one generic evaluative statement and one verbless fragment. This indicates that major meaning errors in MI do not always arise from misrecognition of words. They can also originate in mis-segmentation, which disrupts syntactic dependency and, in turn, deprives the translation module of the structural information required to build a coherent proposition.

**Mis-scoped temporal marker in a contrastive frame**

Source: "Before, a week's work could produce around 11. Now, that takes less than an hour."

System E ASR: "the task of producing bundles for benefits tribunals before a week's work. Could produce around eleven. Now that takes less than an hour."

System E output: "比如在一周的工作前为福利审裁处制作文件包的任务。大约能产出十一个。现在这只需要不到一个小时的时间。"

System B ASR: "before a week's work could produce around eleven. Now that takes less than an hour,"

System B output: "在一周工作量完成之前，大概只能完成十一份。现在这不到一个小时就能完成。"

**Discussion:**

The adverb before here introduces a discourse-level contrast between past and present productivity. Both systems instead attach it locally to the following clause, interpreting it as a temporal modifier within the event structure ("before a week's work is completed") rather than as a contrastive frame marker. The result is a distorted chronology. Human interpreters, by contrast, are more likely to infer the larger comparative frame from the immediate discourse context signalled by Before … Now …, even if the initial adverb is acoustically weak or syntactically detached.

**Supplementary Appendix A. Product-Oriented Codebook**

Coding guide for the evaluation of human and machine English-to-Chinese simultaneous-interpreting output.

This codebook provides the coding guide used in the present study. For scoring purposes, two code groups were included in the main analysis: Meaning and Target-language quality.

Meaning is operationalised as a single code group comprising two sub-code groups—major errors and minor errors—and only these meaning codes contribute to the major-equivalent error count (MEEC).

Target-language quality is assessed separately through an overall score, supplemented by three target-language-quality codes used for qualitative profiling: awkwardness/register, disfluency carryover, and unintelligible or nonsensical output. These codes do not enter the MEEC.

A supplementary process-oriented code, ASR-caused error, was applied only to machine-interpreting outputs in a separate second-round review conducted after the initial blind evaluation was complete and system identities had been revealed. In this second round, the same two raters revisited only the major errors previously coded for machine outputs, comparing each error against the corresponding machine-generated English ASR transcript to determine whether the error originated in ASR failure. This code was used for diagnostic analysis only and never entered the meaning score.

The meaning code group is designed to be exhaustive for the purpose of calculating the major-equivalent error count (MEEC). The target-language quality and cause-of-error code groups, by contrast, are not exhaustive: the codes listed under each were selected for the specific analytical purposes of the present study. Both code groups can accommodate additional codes for studies with different evaluative focuses. More broadly, the coding framework is modular: researchers may select different code groups and individual codes depending on the dimensions and granularity relevant to their research questions. The full coding framework, including a comprehensive account of available code groups and codes and their implementation in ATLAS.ti for interpreting performance evaluation, will be presented in a dedicated methodological paper (Author, in preparation).

**Unit of analysis and coding principles**

Before evaluation, the raters jointly segment each source text into discourse-level key-message segments. Each segment contains one core message together with its supporting language. Meaning codes are assigned with reference to the effect of the rendition on the listener's understanding of the key message of that segment.

Target-language-quality codes may co-occur with meaning codes when both content transfer and target-language acceptability are independently affected.

A segment typically receives at most one major meaning code. A second major code is assigned only when an additional deviation is qualitatively distinct and independently damages the key message. Minor codes are not limited in number within a segment.

**Supplementary Table S1. Code groups used in the study**

| Code group | Codes | Use in the study |
|---|---|---|
| Meaning | 10 | Entered the meaning score (3 major + 7 minor codes) |
| Target-language quality | 3 | Used for target-language-quality profiling and qualitative analysis |
| Cause of error | 1 | Supplementary diagnostic code; applied only to machine outputs in a post-hoc second round (major errors only); never entered the meaning score |

**Meaning codes**

Meaning codes are divided into major and minor errors. Major errors materially damage the key message of a segment; minor errors affect precision, local clarity, or supporting detail without destroying the key message. For aggregate meaning analysis, the major-equivalent error count (MEEC) is calculated as major errors + minor errors / 5.

**Supplementary Table S2. Major meaning codes**

| Code | Operational definition | Quick decision rule | Boundary note |
|---|---|---|---|
| Major meaning/logic error | Substantial distortion of a core proposition, stance, causal/contrast relation, or who-did-what-to-whom relation. | Would a reasonable listener form a wrong understanding of a key point? If yes, code Major. | Use when the key message fails to come across or comes across incorrectly, even if some facts are still |
| | | | present in the output. |
| Major addition | Insertion of substantial content not supported by the source, such that the takeaway, claim, intention, or logic is changed. | Does the added material introduce a new claim or materially redirect the interpretation? If yes, code Major. | Do not use for harmless discourse smoothers (e.g., therefore, in short) unless they impose a new logical relation. |
| Major omission | Loss of information essential to the message, such as the main claim, a key qualifier, a major numerical anchor, or a central action/policy point. | If the omission leaves the message seriously misleading or incomplete, code Major. | If only secondary detail is missing while the main claim remains intact, use Minor omission instead. |

**Supplementary Table S3. Minor meaning codes**

| Code | Operational definition | Quick decision rule | Boundary note |
|---|---|---|---|
| Minor meaning shift | Local inaccuracy that alters nuance, scope, emphasis, or degree while | Can the listener still recover the intended gist | If negation, causality, agency, or the core |
| | preserving the gist of the segment. | with little effort? If yes, code Minor. | proposition is altered, upgrade to Major meaning/logic error. |
| Minor unclear logic | Logical relation is under-specified or locally ambiguous (e.g., connector, comparison, reference) but remains inferable. | Is the relation unclear but recoverable? If yes, code Minor. | If the ambiguity leads the listener toward a wrong conclusion, upgrade to Major meaning/logic error. |
| Minor addition | Small unsupported insertion that does not materially change the meaning but reduces fidelity. | Does the extra content leave the main take-away unchanged? If yes, code Minor. | If the added material changes the claim, condition, or logic, upgrade to Major addition. |
| Minor omission | Loss of non-core detail, such as a secondary qualifier, example, or | Does the main claim remain intact while the local details are lost? If yes, code Minor. | If the omitted item anchors the claim or is essential to the segment's message, |
| | local support, without changing the main claim. | | upgrade to Major omission. If the omitted detail is an item in a list, use Minor list omission instead. |
| Minor list omission | A subtype of minor omission, coded separately so that omissions of list items can be counted on their own. Loss of one or more items in a serial list, parallel structure, or enumeration while the overall function of the list remains recoverable. | Are individual listed items missing, but the broader point of the list still comes through? If yes, code Minor list omission. | If the omitted list item itself carries the key message or changes the action/policy meaning, upgrade to Major omission. An omission coded as Minor list omission is not also coded as Minor omission. |
| Minor terminology | Term choice is understandable but nonstandard, less precise, or mildly inconsistent | Would the listener still understand the intended concept with only minor | If the term changes the core referent or proposition, code Major |
| | with the source-domain meaning. | imprecision? If yes, code Minor. | meaning/logic error; if an essential item is dropped, code Major omission; if the term remains broadly understandable but imprecise, code Minor terminology. |
| Minor number | Small numerical imprecision (e.g., rounding or slight mismatch) that does not alter the overall interpretation. | Would the number error leave the overall claim or decision unchanged? If yes, code Minor. | If order of magnitude, direction, or a critical number is wrong or missing, code Major meaning/logic error or Major omission. |

**Target-language-quality codes**

Target-language-quality codes capture the acceptability and usability of the Chinese output itself. These codes may co-occur with meaning codes when both content transfer and target-language form are independently affected.

**Supplementary Table S4. Target-language-quality codes**

| Code | Operational definition | Quick decision rule | Boundary note |
|---|---|---|---|
| Awkwardness/register | Non-idiomatic, calque-like, or clunky target-language phrasing, or a register mismatch (e.g., inappropriately formal, colloquial, or stylistically inconsistent output), that remains comprehensible. | Understandable but clearly unnatural or register-inappropriate? If yes, code Awkwardness/register. | Do not use this code to judge accent, voice, or delivery; it applies only to target-language form. |
| Disfluency carryover | Target-language output unnecessarily retains source-speech disfluencies, such as fillers, repetitions, false starts, hesitations, or other semantically low-value spoken material, | If the problem is visible in the transcript/output itself, code Disfluency carryover. | Do not confuse this code with delivery fluency. If the problem is mainly in the manner of rendition and requires audio to |
| | instead of filtering or smoothing them. | | judge, it belongs to the delivery dimension, not here. |
| Unintelligible or nonsensical output | The target-language segment becomes hard to parse or semantically incoherent, so that the intended meaning cannot be reliably recovered from the Chinese output alone. | If you cannot confidently paraphrase the Chinese segment, code this category. | If the core problem is omission or distortion, code that first; add this code when target-language breakdown is independently severe. |

**Cause-of-error code**

The following code is supplementary and diagnostic. It is applied only to machine-interpreting outputs and only in a second-round review, after the initial blind evaluation is complete and system identities have been revealed. In this second round, the raters revisit only the major errors previously coded for machine outputs, comparing each error against the corresponding machine-generated English ASR transcript to assess whether the error originated in ASR failure. The judgement rests on what the two columns show: an error is coded as ASR-caused when the Chinese output follows an error visible in the displayed English transcript, such as a misrecognised word or a misplaced sentence boundary. This points to speech recognition as the source of the error. It cannot show that the displayed transcript was the text each system translated from, because the study observed only the on-screen output of each system and did not verify its internal design (see main text, §2.2). Because the first-round evaluation is conducted blind to interpreter identity, raters cannot access ASR transcripts or apply this code during that phase. The code never enters the meaning score.

**Supplementary Table S5. Supplementary cause-of-error code**

| Code | Operational definition | Quick decision rule | Boundary note |
|---|---|---|---|
| ASR-caused error | Applied only to machine-interpreting outputs, in a post-hoc second-round review. After the blind first-round evaluation is complete, raters revisit major errors coded for machine outputs and compare each against the machine-generated English ASR transcript. The code is assigned when the raters judge that | For each major error coded on a machine output: compare the error against the machine-generated ASR transcript. Would the target-side error likely disappear if the recognised source text were correct? If yes, and causality is reasonably clear, apply the code. | This code does not indicate mere co-presence of an ASR anomaly and an output error. Do not apply it when causality cannot be reasonably established. |
| | the observed major error originates in ASR failure rather than in downstream mistranslation alone. | | |

**Severity checklist**

Does the deviation change the core proposition, main stance, major causal/contrast relation, or central action/policy point? If yes, code Major. Would a reasonable listener likely draw a wrong conclusion or miss the key point of the segment? If yes, code Major. Is the problem recoverable with low effort and confined to local precision, support, clarity, or secondary detail? If yes, it is usually Minor. Is the problem primarily one of target-language acceptability rather than content transfer? Apply the relevant target-language-quality code, and add a meaning code as well if the content is also affected.

**Scoring note.** For aggregate meaning analysis, MEEC = number of major errors + number of minor errors / 5. Target-language-quality codes and the ASR-caused error code do not enter the MEEC.

**Implementation note**. All coding was performed in Atlas.ti, with the codes listed above configured as a preset codebook. During evaluation, raters applied codes by selecting the relevant span of the aligned transcript and assigning the preset code directly, ensuring consistent labelling across outputs. After coding was complete, Atlas.ti's built-in export function generated a spreadsheet report of code frequencies per output, from which MEEC and code-level profiles were computed. For studies that require only aggregate meaning accuracy without code-level granularity, the taxonomy can be simplified to two codes (major error and minor error), with the MEEC formula unchanged.

## Supplementary Appendix B. Target-Language Quality Rating Scale

*Adapted from Lee (2008, pp. 172–173)*

The following scale was used to holistically assess target-language quality in interpreting outputs. The original six-band scale from Lee (2008) has been mapped to a 100-point scoring system, with each band serving as an anchor point. A seventh band (score 30, *Very poor*) was added to accommodate outputs that are technically present but largely unintelligible—a performance level not anticipated in the original human-oriented scale but observed in some machine-interpreting outputs. Raters assigned each output to the band that best described its overall target-language acceptability, then recorded the corresponding score. Where performance fell between two anchor points, raters could assign intermediate values (e.g., 75 for performance between *Good* and *Very good*).

| Score | Band | Proficiency level | Descriptor |
|---|---|---|---|
| **90** | 6 | Excellent | Excellent target-language production with few linguistic errors and appropriate target-language expressions. |
| **80** | 5 | Very good | Very good target-language production with a few minor linguistic errors that do not hinder immediate target-language comprehension and generally appropriate target-language expressions. |
| **70** | 4 | Good | Good target-language production with a few linguistic errors that may hinder immediate comprehension, but quite understandable. A few minor inappropriate target-language expressions. |
| **60** | 3 | Adequate | Adequate target-language production with some linguistic errors that hinder comprehension and some inappropriate target-language expressions. |
| **50** | 2 | Inadequate | Inadequate target-language production with many linguistic errors and inappropriate target-language expressions consistently found. |
| **40** | 1 | Poor | Poor target-language production with inappropriate target-language expressions. |
| **30** | — | Very poor | Largely unintelligible or nonsensical target-language production with pervasive linguistic errors that consistently prevent comprehension. Output is technically present but fails to communicate the source message in a recoverable form. |
| **0** | 0 | — | Test abandoned/unfinished; no output produced. |

**Note.** Examples of deviations from target-language norms identified in Lee (2008) include incorrect grammar, unidiomatic language, interference from the source language, and inappropriate register for the target audience and context. The present study also treated disfluencies carried over into the output (repetitions, fillers, false starts) as target-language deviations; Lee lists such features under delivery. Delivery-related criteria (pronunciation, accent, prosody) were not assessed.